\documentclass[a4paper,fleqn]{cas-dc}

\usepackage[numbers]{natbib}
\usepackage{float}
\usepackage{algorithm}
\usepackage{algpseudocode}
\usepackage{graphics} 
\usepackage{epsfig} 
\usepackage{amsmath}
\usepackage{amsfonts}
\usepackage{amssymb}
\usepackage{booktabs}
\usepackage{url}
\usepackage{multirow}
\usepackage{ulem}
\usepackage{bbding}
\usepackage{hyperref}
\usepackage{blindtext}

\def\tsc#1{\csdef{#1}{\textsc{\lowercase{#1}}\xspace}}
\tsc{WGM}
\tsc{QE}
\tsc{EP}
\tsc{PMS}
\tsc{BEC}
\tsc{DE}

\begin{document}
\let\WriteBookmarks\relax
\def\floatpagepagefraction{1}
\def\textpagefraction{.001}

\shorttitle{Referring Flexible Image Restoration}

\shortauthors{Runwei Guan et~al.}

\title [mode = title]{Referring Flexible Image Restoration}                      


\author[1,2,3,4]{Runwei Guan}[style=chinese, orcid=0000-0003-4013-2107]
\fnmark[1]
\ead{runwei.guan@liverpool.ac.uk}
\credit{Conception, Methodology, Experiments, Paper Writing}

\author[5]{Rongsheng Hu}[style=chinese]
\fnmark[1]
\ead{1033170432@stu.jiangnan.edu.cn}
\credit{Dataset Construction, Methodology}

\author[6]{Zhuhao Zhou}[style=chinese]
\ead{xxxak999@gmail.com}
\credit{Dataset Construction}

\author[1,2,3,4]{Tianlang Xue}[style=chinese]
\ead{tianlang.xue22@student.xjtlu.edu.cn}
\credit{Dataset Construction}

\author[3,4]{Ka Lok Man}[orcid=0000-0002-5787-4716]
\ead{ka.man@xjtlu.edu.cn}
\credit{Supervision, Hardware Support}

\author%
[2]
{Jeremy Smith}
\ead{J.S.Smith@liverpool.ac.uk}
\credit{Supervision}

\author%
[4]
{Eng Gee Lim}
\ead{enggee.lim@xjtlu.edu.cn}
\credit{Supervision}

\author%
[6]
{Weiping Ding}[style=chinese, orcid=0000-0002-3180-7347]
\ead{dwp9988@163.com}
\credit{Paper Writing and Correction}

\author%
[1,3,5]
{Yutao Yue}[style=chinese, orcid=0000-0003-4532-0924]
\ead{yueyutao@idpt.org}
\cormark[1]
\credit{Supervision, Hardware Support}

\affiliation[1]{organization={Institute of Deep Perception Technology, JITRI},
    city={Wuxi},
    postcode={214000}, 
    country={China}}

\affiliation[2]{organization={Faculty of Science and Engineering, University of Liverpool,},
    city={Liverpool},
    postcode={L69 3BX}, 
    country={United Kingdom}}

\affiliation[3]{organization={XJTLU-JITRI Academy of Technology, Xi'an Jiaotong-Liverpool University},
    city={Suzhou},
    postcode={215123}, 
    country={China}}

\affiliation[4]{organization={School of Advanced Technology, Xi'an Jiaotong-Liverpool University},
    city={Suzhou},
    postcode={215123}, 
    country={China}}

\affiliation[5]{organization={School of Artificial Intelligence and Computer Science, Jiangnan University},
    city={Wuxi},
    postcode={214122}, 
    country={China}}

\affiliation[6]{organization={School of Electronics and Computer Science, University of Southampton},
    city={Southampton},
    postcode={SO17 1BJ}, 
    country={United Kingdom}}

\affiliation[5]{organization={Thrust of Artificial Intelligence and Thrust of Intelligent Transportation, The Hong Kong University of Science and Technology (Guangzhou)},
    city={Guangzhou},
    postcode={511400}, 
    country={China}}

\affiliation[6]{organization={School of Information Science and Technology, Nantong University},
    city={Nantong},
    postcode={226019}, 
    country={China}}

\cortext[cor1]{Corresponding author: yutaoyue@hkust-gz.edu.cn}



\begin{abstract}
In reality, images often exhibit multiple degradations, such as rain and fog at night (triple degradations). However, in many cases, individuals may not want to remove all degradations, for instance, a blurry lens revealing a beautiful snowy landscape (double degradations). In such scenarios, people may only desire to deblur. These situations and requirements shed light on a new challenge in image restoration, where a model must perceive and remove specific degradation types specified by human commands in images with multiple degradations. We term this task Referring Flexible Image Restoration (RFIR). To address this, we first construct a large-scale synthetic dataset called RFIR, comprising 153,423 samples with the degraded image, text prompt for specific degradation removal and restored image. RFIR consists of five basic degradation types: blur, rain, haze, low light and snow while six main sub-categories are included for varying degrees of degradation removal. To tackle the challenge, we propose a novel transformer-based multi-task model named TransRFIR, which simultaneously perceives degradation types in the degraded image and removes specific degradation upon text prompt. TransRFIR is based on two devised attention modules, Multi-Head Agent Self-Attention (MHASA) and Multi-Head Agent Cross Attention (MHACA), where MHASA and MHACA introduce the agent token and reach the linear complexity, achieving lower computation cost than vanilla self-attention and cross-attention and obtaining competitive performances. Our TransRFIR achieves state-of-the-art performances compared with other counterparts and is proved as an effective architecture for image restoration. We release our project at \url{https://github.com/GuanRunwei/FIR-CP}.
\end{abstract}




\maketitle

\section{Introduction}
Based on the continuous advancements in deep learning in recent years, there are two popular prevalent paradigms in image restoration. The first one involves task-agnostic models for image restoration \cite{liang2021swinir,zamir2022restormer, li2023efficient}, trained on datasets with diverse degradations separately \cite{wang2022uformer,zhang2023unified}, exhibiting effective performance across various degradations \cite{chen2023ipdnet,chen2022simple}. However, this paradigm is still specific to particular degradation, with a model structure ensuring good generalization in various degradation domains. The second paradigm named the all-in-one model adopts a learnable degradation encoding approach \cite{chen2021pre,guo2023adaptir,li2022all}, capable of adaptively learning features associated with different degradations \cite{chen2023always,li2020all,potlapalli2023promptir,valanarasu2022transweather} and encoding them in a latent space to guide the decoder for image restoration \cite{wang2023gridformer,zhu2023learning,ozdenizci2023restoring,xu2024unified,liu2022tape}. This approach embeds knowledge about various interference types into a single model. For this model, the learnable degradation encoding encompasses both implicit and explicit knowledge, yielding impressive restoration results.

\begin{figure}
  \includegraphics[width=0.49\textwidth]{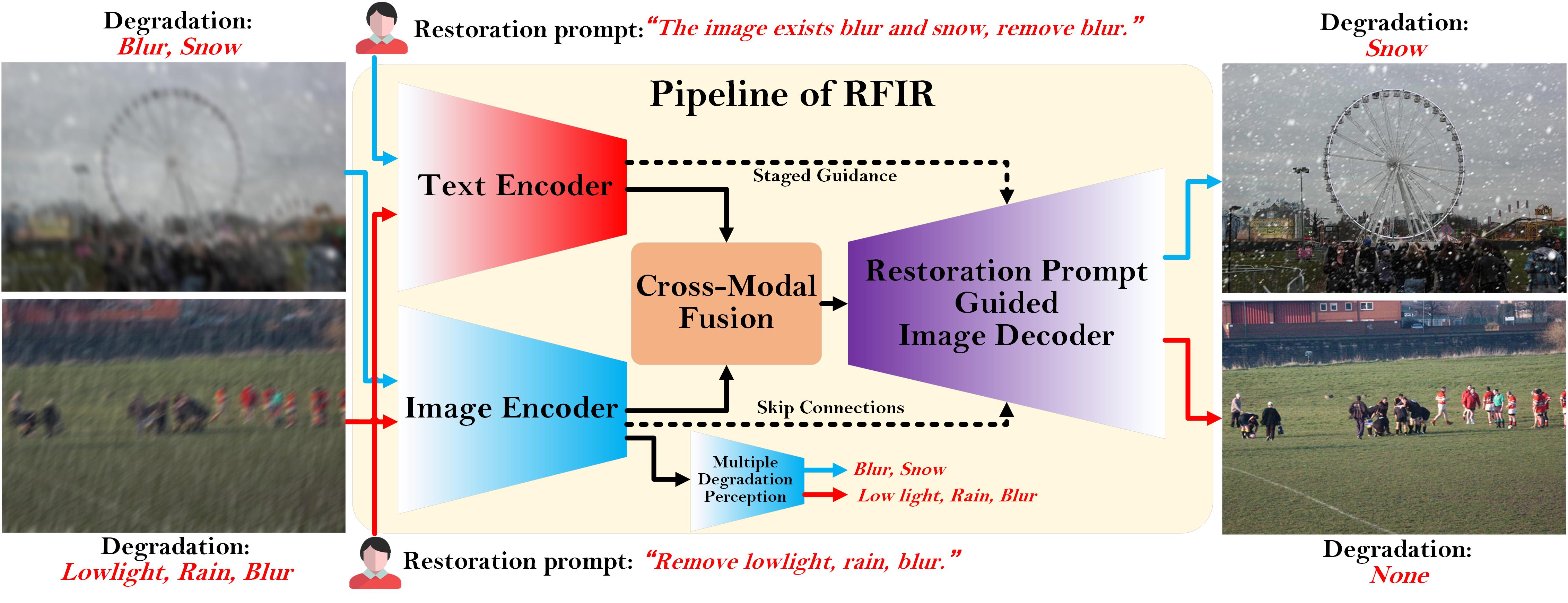}
  \vspace{-4mm}
  \caption{\textmd{The overview of our proposed pipeline, including input image with degradations, text-guided restoration model and predictions. Two samples (a degraded image, a restoration prompt and a restored image), contains the partial degradation removal (upper) and global degradation removal (nether).}}
  \label{fig:teaser}
\end{figure}

While models of the second paradigm can identify different degradations in distinct images, this is solely based on the assumption that one image contains only one type of degradation, rendering them inefficient in recognizing multiple degradations within a single image. In real-world scenarios, multiple degradations often coexist within the same image. For instance, capturing a photo with a smartphone lens on a night with heavy rain may result in a blurry image, encompassing three types of degradation: low light, rain, and blurriness. In such cases, users tend to prefer removing all degradation simultaneously. Another instance is the lens in front of a crop spraying system experiencing haze; in reality, there is only one type of degradation, which is haze. The water sprayed by the system is not considered degradation, so the observer only wants to eliminate the haze. However, task-specific models like Restormer \cite{zamir2022restormer} can only remove the specific degradation. Besides, all-in-one models like PromptIR \cite{potlapalli2023promptir} are capable of performing image restoration to a certain extent under multiple degradations, but the results are not very satisfactory. Furthermore, whatever task-specific or all-in-one models, cannot perceive human intentions and are tempted to remove all possible degradations, making them incapable of image restoration in response to specific user preferences.


\begin{figure*}
    \includegraphics[width=0.99\linewidth]{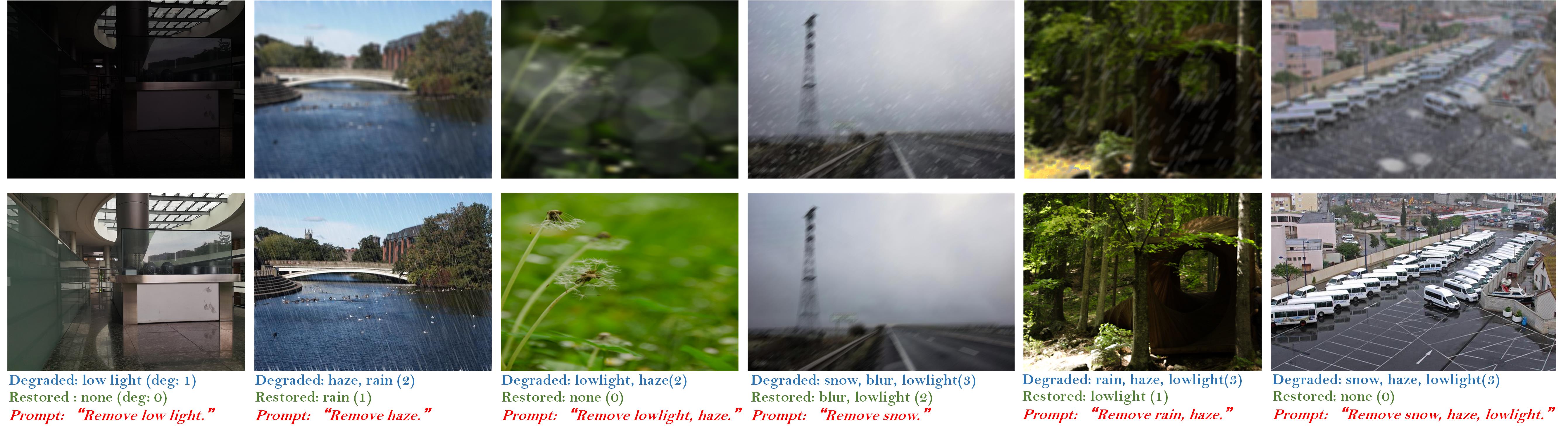}
    \vspace{-2mm}
    \caption{\textmd{Samples of RFIR, including six categories (Table \ref{tab:rfir}) of referring specific degradation removal. Each sample contains the degraded image (the first row), restored image (the second row) and text prompt to remove specific degradation(s).}}
    \label{fig:example_samples}
\end{figure*}

Recently, deep neural network paradigms based on Image and Language Fusion (ILF) have made impressive progress in computer vision over the past few years \cite{uppal2022multimodal}. One of the crucial reasons is that natural language can serve as an effective supervisory signal to control deep learning models to sample and map image features for downstream tasks flexibly \cite{zhou2022learning}. Although most ILF works are oriented to high-level image tasks, such as visual grounding \cite{qiao2020referring}, ILF-based models for low-level tasks such as image restoration and super-resolution gradually start to show their superior generalization and user-friendliness \cite{liang2023iterative,qi2023tip,chen2023image,yu2024scaling}, which also reveal competitive performances compared with task-specific models, general restoration models and all-in-one restoration models. Nevertheless, we notice that these efforts are currently centered around using natural language to guide the restoration of images affected by a single type of degradation, achieving notable generalization. Nevertheless, there is currently a lack of research focusing on the flexible restoration of specific degradation in images with multiple degradation contexts using natural language. This field remains unexplored at present and has several challenges. We initially partition the entire task into three stages: perception, fusion, and restoration.

\begin{enumerate}
    \item For a given RGB degraded image, multiple degradations may result in feature overlap, and the same type of degradation exhibits intra-domain severity. In the case of compressed features, the model needs to adaptively separate and model different degradations in anisotropic images within a high-dimensional feature latent space.
    \item Multi-head cross-attention is capable of dynamically globally modeling and fusing features from heterogeneous modalities. However, its quadratic complexity makes it computationally expensive. Hence, there is a need to design a more cost-effective, lightweight cross-attention mechanism.
    \item During the image restoration process, it is crucial to prevent the model from forgetting fusion knowledge in multiple decoding stages. This ensures that the textual prompt continuously aligns and guides the removal of specific degradations from the image features.
\end{enumerate}

Therefore, in this paper, we focus on the removal of specified degradation and image restoration for images with multiple degradations, utilizing natural language signals. Our contributions are listed as follows,

\begin{enumerate}
    \item We propose a new challenging task in image restoration called Referring Flexible Image Restoration, which is aimed to remove specific degradations and restore images according to given text prompts. It tackles limitations where individuals are unable to control the degree of image restoration according to their own intentions.
    \item We make and establish the first large-scale dataset for flexible image restoration based on natural language, called Referring Flexible Image Restoration (RFIR). RFIR focuses on flexible image restoration under adverse conditions including blur, rain, haze, snow, and low light, where single degradation as well as combinations of double and triple degradations are included. It comprises 153,423 samples, each consisting of a degraded image, ground truth image, and corresponding textual prompts.
    \item We propose a novel end-to-end multi-task model called TransRFIR for referring flexible image restoration, which is capable of simultaneously perceiving different types of degradation present in an image and effectively removing specific degradation guided by natural language prompts. Moreover, we propose a lightweight and effective cross attention module called Multi-Head Agent Cross Attention (MHACA) to fuse features of image and text with linear complexity. TransRFIR achieves state-of-the-art performances compared with other counterparts.
    \item Our proposed paradigm in TransRFIR (Figure \ref{fig:teaser}) can be migrated elegantly to other U-Net-based image restoration networks to guide image features for nicely flexible image restoration.
\end{enumerate}

The remaining content is organized as follows, Section \ref{sec:related} states the related works regarding image restoration for multi-degradations and natural language driven image restoration; Section \ref{sec:dataset} illustrates our proposed RFIR dataset, inlcuding overall statistics and data construction; Section \ref{sec:method} demonstrates the methods, containing our proposed pipeline and two attention modules; Section \ref{sec:experiments} is the comprehensive experiments; Section \ref{sec:discussion} discusses the specific motivation (a little story), limitations and challenges of our work.

\begin{table*}
    \centering
  \setlength\tabcolsep{0.5pt}
  \caption{\textmd{Statistics of RFIR Dataset for degradations.}}
  \vspace{-1mm}
  \label{tab:rfir}
  \begin{tabular}{c|ccc|cccc|ccccc|c}
    \toprule
    \multirow{2}[2]{*}{\textbf{Datasets}} & \multicolumn{12}{c}{\textbf{Sample Types}} \vline & \multirow{2}[2]{*}{\textbf{Total Sum}}\\
    \cmidrule(lr){2-13}
     & \textcolor{red}{One}-\textcolor{blue}{One} & \textbf{Sum} & \textbf{Pct(\%)} & \textcolor{red}{Two}-\textcolor{blue}{One} & \textcolor{red}{Two}-\textcolor{blue}{Two} & \textbf{Sum} & \textbf{Pct(\%)} & \textcolor{red}{Three}-\textcolor{blue}{One} & \textcolor{red}{Three}-\textcolor{blue}{Two} & \textcolor{red}{Three}-\textcolor{blue}{Three} & \textbf{Sum} & \textbf{Pct(\%)} \\
    \midrule
    Train & 51164 & 51164 & & 32694 & 8178 & 40872 & & 6135 & 6129 & 3096 & 15360 & & \textbf{107396} \\
    Validation & 7451 & 7451 & \textbf{47.6} & 4613 & 1131 & 5744 & \textbf{38.1} & 871 & 849 & 427 & 2147 & \textbf{14.3} & \textbf{15342} \\
    Test & 14440 & 14440 & & 9464 & 2364 & 11828 & & 1756 & 1783 & 878 & 4417 & & \textbf{30685} \\
    \bottomrule
  \end{tabular}
  \\
  \footnotesize {\textcolor{red}{Red: Degradation number per image.} \textcolor{blue}{Blue: Degradation number referred by text prompt per image.} }
\end{table*}

\section{Related Works}
\label{sec:related}

\subsection{Image Restoration for Multiple Degradations}
In real-world scenarios, both single and multiple degradations coexist. However, compared to the restoration of images with a single degradation \cite{zamir2022restormer,liang2021swinir,ren2021adaptive,tsai2022banet, cai2023retinexformer,wang2022uformer,li2023efficient}, image restoration for multiple degradations remains relatively unexplored. \cite{liu2022tape,li2020all,li2022all,chen2021pre} design parallel image encoders or decoders for different degradation, where they require prior knowledge of specific degradation types, which limits their scalability. Considering the lack of any prior information about degradation in blind image restoration, \cite{potlapalli2023promptir,chen2023always,luo2023controlling} employ learnable degradation encoding to implicitly learn different degradation patterns present in the images. This allows for the automatic removal of various types of degradation. Besides, \cite{luo2023controlling} constructs a dataset containing multiple degradations for training, but each sample contains one degradation only. Furthermore, current state-of-the-art image restoration models are built upon the transformer architecture. Due to the anisotropy in images with different backgrounds and degradations, multi-head self-attention in transformers can dynamically and adaptively restore images. In contrast, the weights of Convolutional Neural Networks (CNNs) are static, which makes it difficult to match the performance of transformer-based models in most cases. However, vanilla self-attention suffers from quadratic growth in computational complexity, hence how to build an efficient self-attention module with low computational complexity is very important.

\subsection{Natural Language Driven Image Restoration}
Natural language-driven fusion models are gaining popularity in various visual tasks, including localization \cite{wu2023language,guan2024watervg}, captioning \cite{hossain2019comprehensive}, and visual question answering \cite{sima2023drivelm}. However, applying this language-visual fusion paradigm to more fine-level image processing tasks such as deblurring, denoising, rain and fog removal, super-resolution, etc., remains an open research question and is relatively unexplored. \cite{qi2023tip} is a text-driven diffusion model for joint image denoising and editing, where the restoration strength can be quantified by text prompts. \cite{chen2023image} introduces textual prompts to provide degradation priors for enhancing image super-resolution. The textual prompts are embedded by a language model and injected into the diffusion model through cross-attention, guiding the process of super-resolving the image. \cite{luo2023controlling} leverages a large-scale pre-trained visual language model for general image restoration, with textual instructions containing degradation priors and image background captions. \cite{conde2024high} propose a text-guided image restoration model called InstructIR for various degradations.
Based on the above, the aforementioned works focus on single degradation removal and image editing, utilizing large deep learning models, so we identify that the question of driving vision models for the restoration of multiple degradations, especially using a cost-effective lightweight solution remains an unresolved issue.


\section{The Dataset of Referring Flexible Image Restoration}
\label{sec:dataset}
We propose the Referring Flexible Image Restoration (RFIR) dataset, a text prompt guided restoration dataset for specific degradation removal, where each samples contain a degraded image, a restored image, two type degradation removal prompts. We illustrate RFIR in Section \ref{subsec:all_stat} and \ref{subsec:mdtpg}.

\begin{figure}[h]
    \includegraphics[width=0.94\linewidth]{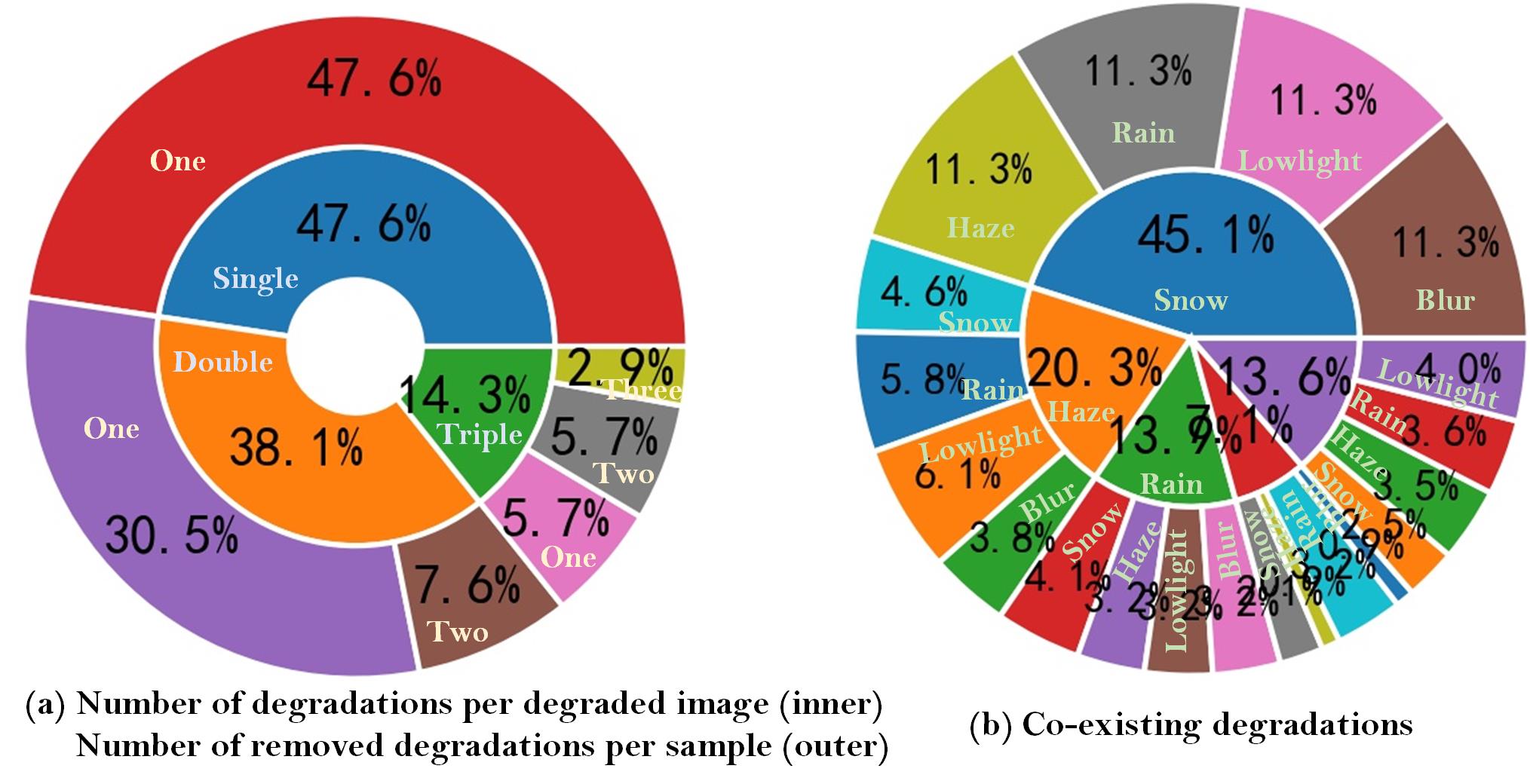}
    \vspace{-3mm}
    \caption{\textmd{(a) Proportion of existing degradations in degraded images (inner) and removed degradations (outer) in restored images; (b) Co-existing degradation types.}}
    \label{fig:dataset_pie}
\end{figure}

\begin{table}[h]
    \setlength\tabcolsep{6.5pt}
    \centering
    \caption{The basic data of RFIR dataset on multiple degradations.}
    \begin{tabular}{c|c}
    \toprule
       \textbf{Degradation Removal}  & \textbf{Datasets}  \\
    \midrule
        Deblur & RIM \cite{cho2021rethinking}, GoPro \cite{nah2017deep} \\
        \midrule
        \multirow{2}[2]{*}{DeHaze} & Haze4K \cite{liu2021synthetic}, Dense-Haze \cite{ancuti2019dense}, \\
        & RESIDE \cite{li2018benchmarking} \\
        \midrule
        Lowlight Enhancement & LOL \cite{Chen2018Retinex}, LSRW \cite{hai2023r2rnet} \\
        \midrule
        DeRain & Rain1400 \cite{fu2017removing}, Rain100H \cite{yang2016joint} \\
        \midrule
        DeSnow & Snow100K \cite{liu2018desnownet} \\
    \bottomrule     
    \end{tabular}
    \label{tab:rfir_base_data}
\end{table}

\subsection{Overall Statistics}
\label{subsec:all_stat}

As shown in Table \ref{tab:rfir_base_data}, RFIR comprises 153,423 samples with single degradation collected from RIM \cite{cho2021rethinking}, RESIDE \cite{li2018benchmarking}, GoPro \cite{nah2017deep}, Haze4K \cite{liu2021synthetic}, Dense-Haze \cite{ancuti2019dense}, LSRW \cite{hai2023r2rnet}, LOL \cite{Chen2018Retinex}, Rain1400 \cite{fu2017removing}, Rain100H \cite{yang2016joint}, Snow100K \cite{liu2018desnownet}, forming the sample pool that includes five types of degradations: blur, haze, low light, rain, and snow. The dataset is divided into training, validation, and test sets, with 107,396, 15,342, and 30,685 samples, respectively. Notably, the dataset consists of images with single degradation accounting for 47.6\%, double degradations for 38.1\%, and triple degradations for 14.3\%. In detail, as illustrated in Table \ref{tab:rfir}, for images with double degradations, there are sub-categories involving the removal of one (partial) or two (global) degradations. For the image with triple degradations, the sub-categories contain the removal of one (partial), two (partial)
or three (global) degradations. Besides, Figure \ref{fig:dataset_pie} (b) presents the proportion of coexistence between degradations. Importantly, single degradation images are original degraded samples from the initial sample pool, while double and triple degradation samples are synthetically generated in later stages, as discussed in detail in Section \ref{subsec:mdtpg}.

\subsection{Multi-Degradation and Prompt Generation}
\label{subsec:mdtpg}

\begin{figure}
    \includegraphics[width=0.99\linewidth]{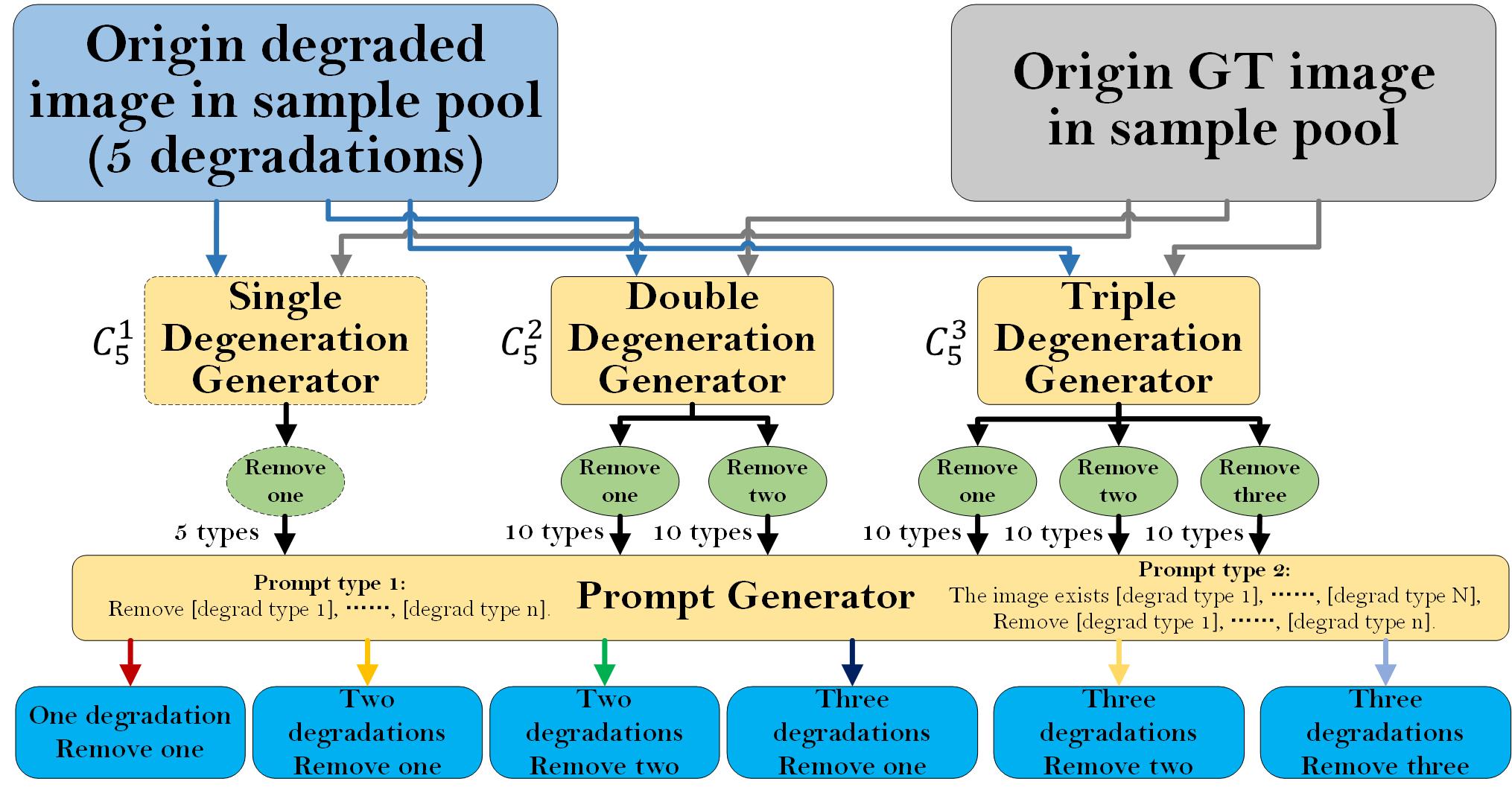}
    \vspace{-4mm}
    \caption{\textmd{Construction process of RFIR dataset.}}
    \label{fig:make_dataset}
\end{figure}

\begin{figure}
    \includegraphics[width=1.00\linewidth]{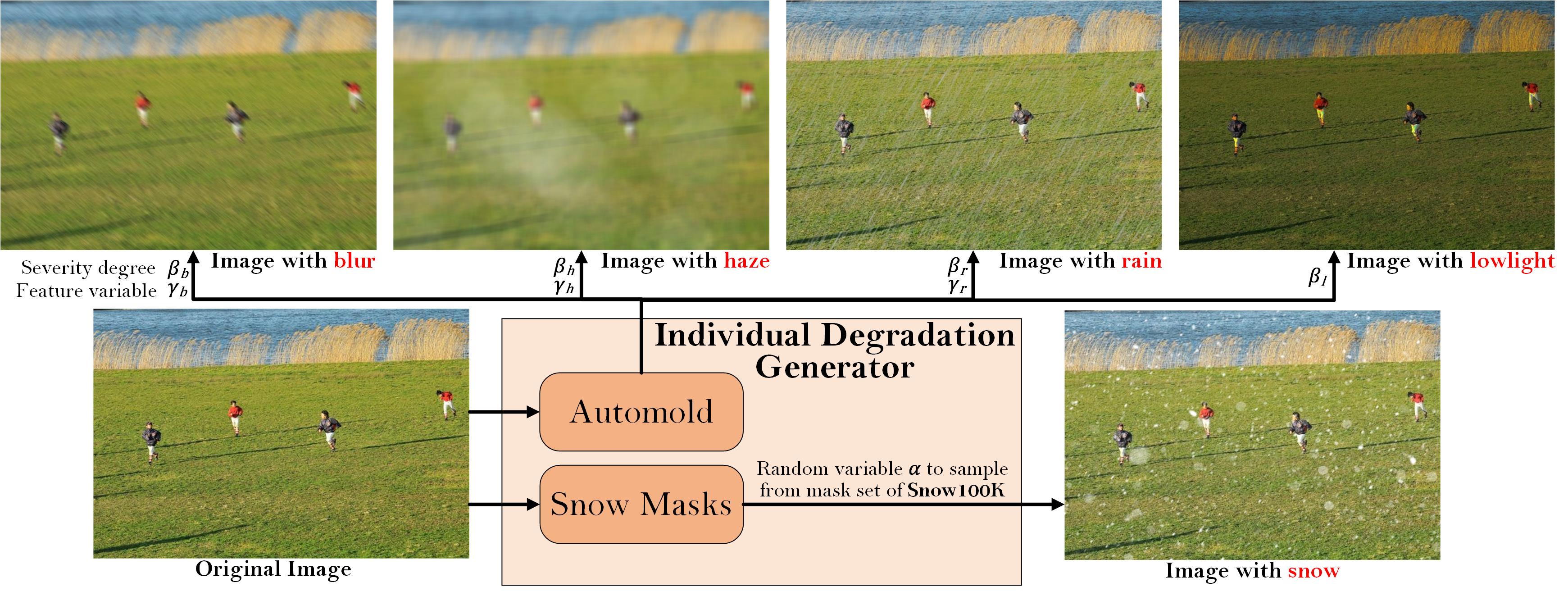}
    \vspace{-4mm}
    \caption{\textmd{Individual degradation generation.}}
    \label{fig:individual_degradation_generator}
\end{figure}

\textbf{Multi-Degradation Generation.} As Figure \ref{fig:make_dataset} presents, based on the images with a single degradation and corresponding ground truth images in the sample pool, we sequentially pass them through the single, double, and triple degradation generators. \textbf{Firstly}, for single degradation, we directly use samples from the original pool. \textbf{Secondly}, as shown in Figure \ref{fig:make_dataset}, the double and triple degradation generators, based on \cite{weather_aug} and \cite{liu2018desnownet}, synthesize additional degradations different from the original single degradation on the images. Exactly, as Figure \ref{fig:individual_degradation_generator} shows, for degradation appending of blur, haze, rain and low light, we adopt the refined degeneration synthesis tools called Automold \cite{weather_aug}. For the snow appending, we adopt snow masks extracted from Snow100K \cite{liu2018desnownet}. For snow generation, the random variable $\alpha$ is for sampling the snow mask from Snow100K's set. For the generation of blur, haze and rain, two variables, $\beta$ and $\gamma$ are to control the severity degree and feature. The feature here represents the degradation itself bringing about some visually distinctive characteristics. For example, variations corresponding to blurred directions due to different motion directions, differences in the positions of regions with fog accumulation, and the degree of tilt of raindrops. For the generation of low light, only one variable $\beta$ is to control the darkness severity. Therefore, our RFIR contains both inter-domain differences of various degradations and intra-domain differences of degradation severity levels. \textbf{Additionally}, if the sample requires the removal of all degradations, the ground truth image in the sample pool serves as the ground truth for that sample. Otherwise, if only specific partial degradations are to be removed, the ground truth sample will be added degradations other than those being removed. It is noteworthy that in the synthesis of multi-degradation images, we ensure that there is no severe degradation feature masking and complete invisibility of image background information among the disturbances. This is done to prevent the model from optimizing in the wrong direction during image restoration. 

\begin{figure}
    \includegraphics[width=0.99\linewidth]{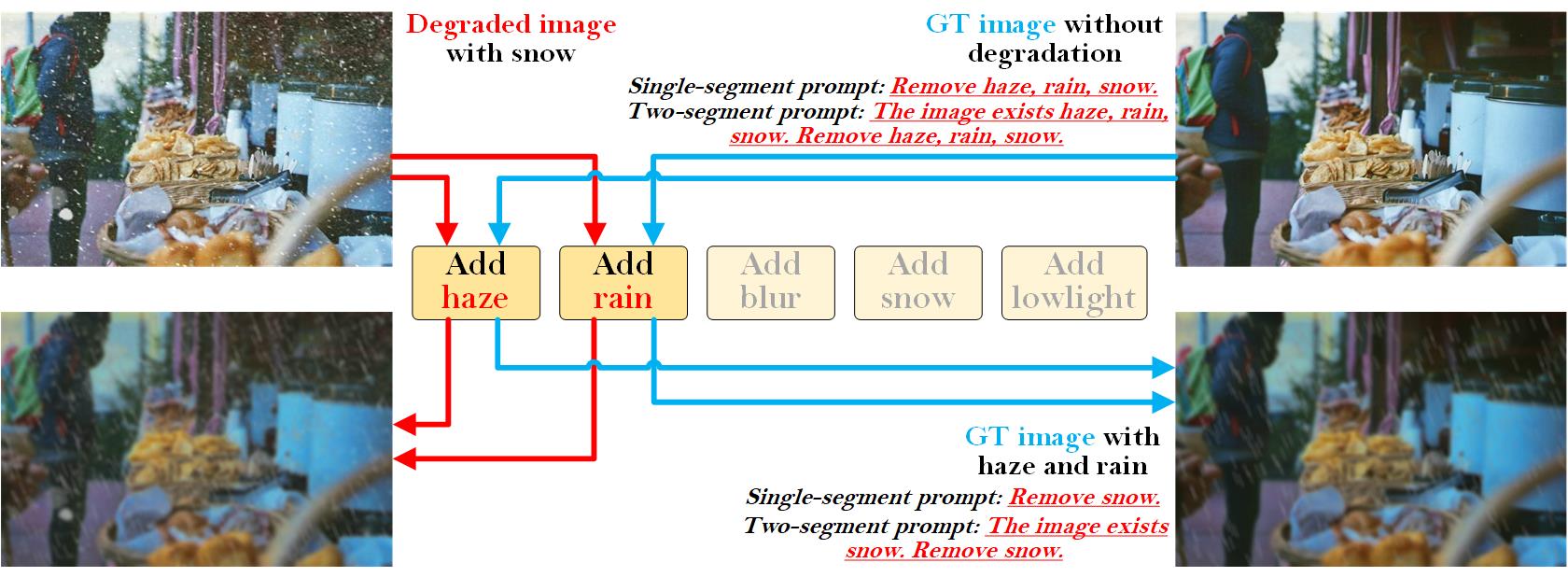}
    \vspace{-4mm}
    \caption{\textmd{Single and Multiple Degradation Generator.}}
    \label{fig:make_dataset_2}
\end{figure}

\textbf{Prompt Generation.} We generate specific text prompts for each sample through the prompt generator. The text prompt comes in two types, as illustrated in Figure \ref{fig:make_dataset_2}. The first type only expresses which specific degradations to remove (single-segment). In contrast, the second type indicates the degradations present in the degraded image (strong prior) and specifies which particular degradations should be removed (two-segment).

\section{Method}
\label{sec:method}

\begin{figure*}
    \includegraphics[width=0.99\linewidth]{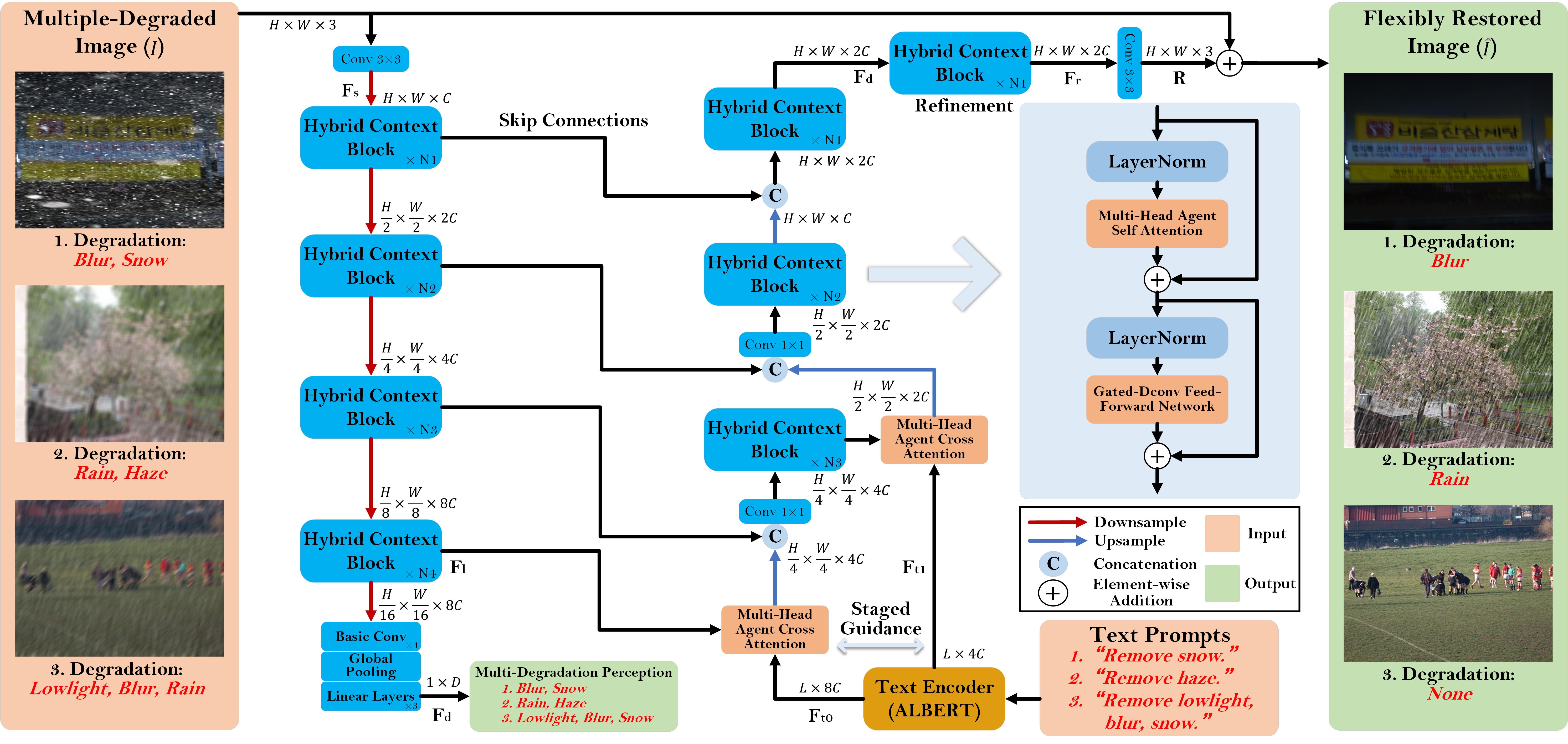}
    \vspace{-3mm}
    \caption{\textmd{The architecture of TransRFIR, contains a four-stage image encoder, text encoder, cross-attention fusion module, four-stage image decoder, and multi-degradation perception classifier. TransRFIR has two inputs (a degraded image and a text prompt) and one output (a restored image where the removed degradations are referred in the text prompt).}}
    \label{fig:transrfir}
\end{figure*}

To efficiently accomplish text-guided image restoration, we propose a multi-task image-text fusion model for image restoration, balancing restoration performance and model simplicity. We elaborate on the following four aspects in detail: \ref{subsec:overall}. Overall Pipeline \ref{subsec:hcblock}; Hybrid Context Block \ref{subsec:mhaca}; Multi-Head Agent Cross Attention \ref{subsec:mdp}; Multiple Degradation Perception \ref{subsec:mdp}; Training Objectives \ref{subsec:training_obj}.

\subsection{Overall Pipeline}
\label{subsec:overall}
Given an image with single or multiple degradations $I \in \mathcal{R}^{H\times W\times 3}$, TransRFIR first adopts a 3$\times$3 convolution on it to obtain the shallow feature $F_s \in \mathcal{R}^{H\times W\times C}$, where $H \times W$ is the spatial size and $C$ is the channel dimension. Subsequently, the shallow feature undergoes a 4-stage image encoder called Hybrid Context Encoder (HCEncoder), to obtain the latent feature $F_l \in \mathcal{R}^{\frac{H}{8}\times \frac{W}{8}\times 8C}$. $F_l$ encompasses a compressed representation of degradation within the image space, as well as background information. In the final stage of HCEncoder, two branches are extended, with one branch further down-sampling based on $F_l$ and entering a multi-degradation perception classifier. The multi-degradation perception classifier outputs a feature $F_d \in \mathcal{R}^{1 \times D}$ representing the probability of multiple degradations in the degraded image, where $D$ is the number of multi-degradation categories.

Furthermore, another branch from the latent feature $F_l$ is fused with the text feature $F_{t}^1 \in \mathcal{R}^{L \times 8C}$ encoded by the text encoder ALBERT \cite{lan2019albert} through the Multi-Head Agent Cross-Attention (MHACA), where $L$ is the embedding length and $8C$ is the channel dimension projected by the feedforward layer. The output of MHACA is consistent with the size of the image latent features $F_l$ and is fed into the Restoration Prompt Guided Image Decoder for staged decoding. To prevent possible knowledge forgetting in multi-stage decoding, we employ MHACA to fuse the visual decoding features in the second stage of the decoder with the features $F_{t}^2 \in \mathcal{R}^{L \times 4C}$ from another feedforward branch extended from the text encoder. Subsequently, the features undergo progressive decoding in the decoder to obtain features $F_d \in \mathcal{R}^{H\times W\times 2C}$ with the same spatial dimensions as the original image. Through a refinement stage, deep features are further enriched to obtain features $F_r \in \mathcal{R}^{H\times W\times 2C}$. $F_r$ undergo adaptive dimension reduction through a $3\times 3$ convolution, resulting in the feature $R \in \mathcal{R}^{H\times W\times 3}$ with the same dimensions as the original image. $R$ is then added to the input image $I$ through a long residual path to obtain the restored image $\hat{I}$, which is guided by natural language for specific degradation removal. Besides, we apply pixel-unshuffle and pixel-shuffle operations on feature downsampling and upsampling, respectively. To enrich deep features and enhance restoration efficiency, we employ skip connections and channel concatenation to connect and fuse features between the encoder and decoder. Channel dimensionality reduction is achieved through a 1×1 convolution. The entire TransRFIR model is based on Agent Attention, a lightweight self-attention with lower complexity that enables fine-grained encoding of global context. We will provide detailed explanations in Section \ref{subsec:hcblock}, \ref{subsec:mhaca}, \ref{subsec:mdp} and \ref{subsec:training_obj}.

\subsection{Hybrid Context Block}
\label{subsec:hcblock}

We propose the Hybrid Context Block (HCBlock) as the foundational module for TransRFIR (Figure \ref{fig:transrfir}). Given the complex context of multiple degradations in the input, where different degradations exhibit distinct visual features, we aim to dynamically encode images with anisotropic degradations using the self-attention-based architecture rather than convolutions with static weights. However, conventional self-attention models exhibit quadratic growth in complexity based on the image size. To reduce model complexity while preserving the capability for global feature modelling, we design the Multi-Head Agent Self-Attention (MHASA) module based on \cite{han2023agent} as a core component of HCBlock. HCBlock follows the architectural design of MetaFormer \cite{yu2022metaformer} and includes LayerNorm, MHASA, Gated-Dconv Feedforward Network (GDFN) \cite{zamir2022restormer}, and two residual connections, where MHASA takes charge of global contextual encoding. GDFN performs redundant filtering and hierarchical enrichment of features upon gated mechanism.

\begin{figure}[h]
    \includegraphics[width=0.99\linewidth]{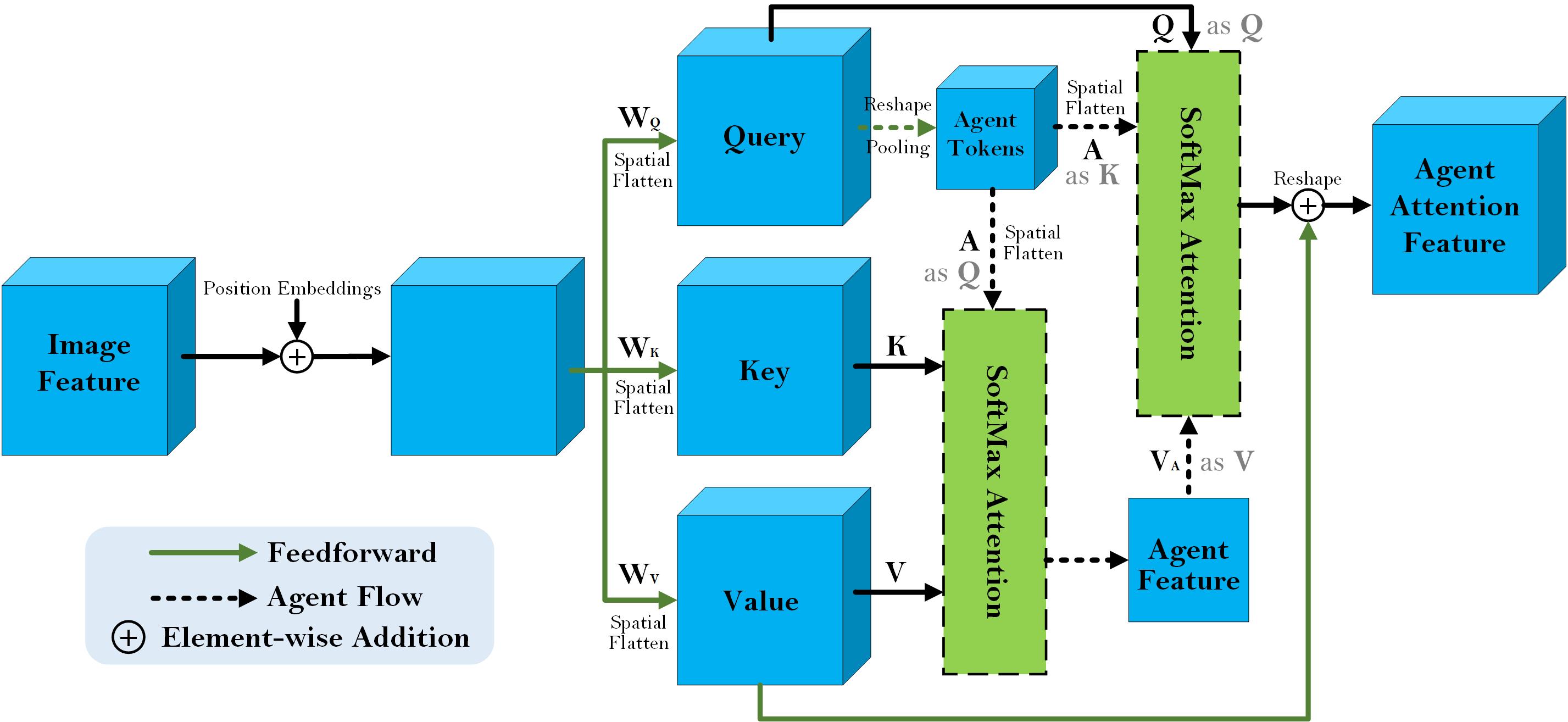}
    \vspace{-3mm}
    \caption{\textmd{The architecture of Agent Self-Attention.}}
    \label{fig:agent_attn}
\end{figure}

As Figure \ref{fig:agent_attn} presents, we define the input feature of MHASA as $F \in \mathcal{R}^{H\times W\times C}$, we first add the learnable position encoding \cite{dosovitskiy2020image} to $F$ and get $F_p \in \mathcal{R}^{H\times W\times C}$. $Q, K, V  \in \mathcal{R}^{N \times C}$ are obtained by the feedforward projection (Equation \ref{eq:qkv1}, \ref{eq:qkv2} and \ref{eq:qkv3}) and dimension flattening on $F_p$.

\begin{align}
    Q & = Flat(F_p W_Q), Q \in \mathcal{R}^{N \times C}, \label{eq:qkv1}\\
    K & = Flat(F_p W_K), K \in \mathcal{R}^{N \times C}, \label{eq:qkv2}\\
    V & = Flat(F_p W_V), V \in \mathcal{R}^{N \times C}, \label{eq:qkv3}
\end{align}

where $W_Q$, $W_K$ and $W_V$ denote feedforward projection matrices for $Q$, $K$ and $V$, where $N=H \times W$.

We formulate softmax attention as Equation \ref{eq:softmax_attn},

\begin{equation}
    \sigma(Q, K, V) = \frac{Q K^T}{\sqrt{d}} \cdot V,
    \label{eq:softmax_attn}
\end{equation}
where $\sigma$ denotes softmax attention while $W_d$ denotes the feedforward function. $d$ denotes the dimension of $Q$. 

We initially treat the agent token $A \in \mathcal{R}^{H_AW_A\times C}$ as an agent query of $Q$ and perform softmax attention calculation (Equation \ref{eq:softmax_attn}) between $K$ and $V$ to aggregate the agent feature $V_A$ from all values, where $H_AW_A = n$. Subsequently, in the second softmax attention with the query $Q$, we use $A$ as the key and $V_A$ as the value, broadcasting the global feature from the agent feature to each query token, resulting in the output feature $\hat{F}$. To enhance the feature diversity, we follow \cite{han2023flatten} to apply a feedforward module by depth-wise convolution added to the $\hat{F}$ and reshape it to the image-like feature. The whole process is presented in Equations \ref{eq:mhasa1}, \ref{eq:mhasa2} and \ref{eq:mhasa3}.

\begin{align}
    V_A & = \sigma(A, K^T, V), V_A \in \mathcal{R}^{n \times C}, 
    \label{eq:mhasa1}
    \\
    \hat{F} & = \sigma (Q, A^T, V_A) + VW_{d}, \hat{F} \in \mathcal{R}^{N \times C}, 
    \label{eq:mhasa2}
    \\
    \hat{F} & = Reshape(\hat{F}),  \hat{F} \in \mathcal{R}^{H \times W \times C},
    \label{eq:mhasa3}
\end{align}

Hence we avoid the direct pairwise softmax attention calculation between $Q$ and $K$ while maintaining the feature interaction between $Q$ and $K$ through the agent $A$.

Here, we define the sequence length of $Q$, $K$ and $V$ as $N=H \times W$ while the sequence length of $A$ as $n = H_A \times W_A$, where $n$ is usually set to a small number in practical and $n < N$ and $n < d$ in the normal situations within neural networks. $d$ is the channel dimension of the feature. Therefore, the computation complexity of agent self-attention is $O(Nnd)$, which is smaller than MDTA's $O(Nd^2)$ \cite{zamir2022restormer} and vanilla self-attention's $O(N^2d)$ \cite{dosovitskiy2020image}.

\subsection{Multi-Head Agent Cross Attention}
\label{subsec:mhaca}

\begin{figure}[h]
    \includegraphics[width=0.99\linewidth]{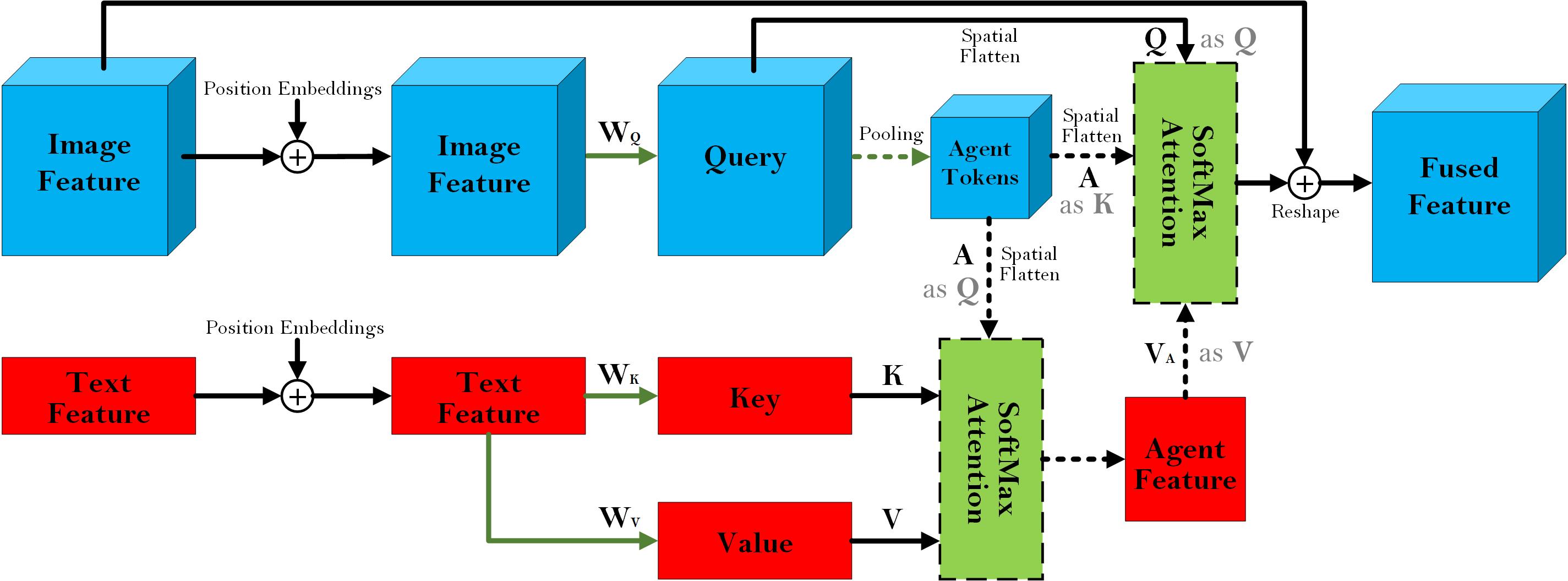}
    \vspace{-3mm}
    \caption{\textmd{The architecture of Agent Cross-Attention.}}
    \label{fig:agent_cross_attn}
\end{figure}

Drawing inspiration from \cite{han2023agent}, we propose a high-efficiency image-text cross-attention module called Multi-Head Agent Cross Attention (MHACA). Compared to vanilla Multi-Head Cross Attention \cite{dosovitskiy2020image} and Multi-Head Linear Cross Attention (MHLCA) \cite{choromanski2020rethinking}, MHACA exhibits lower computational complexity and resources while efficiently fusing heterogeneous features, degraded images and text prompts specifying the removal of particular degradations.

As Figure \ref{fig:agent_cross_attn} presents, we first project the image and text features to the same channel dimension, where we obtain a pair of inputs, the image feature $F_I \in \mathcal{R}^{H \times W \times C}$ and the text feature $F_T \in \mathcal{R}^{L \times C}$. $L$ is the length of text feature. Next, we add corresponding positional encodings $P_I$ and  $P_T$ to both \cite{dosovitskiy2020image}\cite{vaswani2017attention}, which serve to identify the spatial positions of features and alleviate potential loss of positional information arising from dimension transformations. Subsequently, the image and text features propagate forward to obtain $Q \in \mathcal{R}^{H \times W \times C}$, and $K, V \in \mathcal{R}^{L \times C}$. The whole process is shown in Equations \ref{eq:cross_attn_initial_0}, \ref{eq:cross_attn_initial_1} and \ref{eq:cross_attn_initial_2}.

\begin{align}
    & Q = F_I W_Q + P_I, Q \in \mathcal{R}^{H \times W \times C} 
    \label{eq:cross_attn_initial_0} \\
    & K = F_T W_K + P_T,  K, V \in \mathcal{R}^{L \times C}
    \label{eq:cross_attn_initial_1} \\
    & V = F_T W_V + P_T,  
    \label{eq:cross_attn_initial_2}
\end{align}

As Equation \ref{eq:cross_attn1}, \ref{eq:cross_attn2}, \ref{eq:cross_attn3} and \ref{eq:cross_attn4} present, we first exert adaptive pooling on $Q$ and obtain the agent token $A \in \mathcal{R}^{H_A \times W_A \times C}$, which maintains the overall spatial context of $Q$. Then the flattened $A \in \mathcal{R}^{H_AW_A (d) \times C}$ is to perform softmax attention operation with $K$ and obtain the similarity matrix, which is for the calculation of the agent attention feature with $V$. Meanwhile, the agent feature is broadcast by the second softmax attention with flattened $Q \in \mathcal{R}^{HW (N) \times C}$ and $A$ to obtain the image-conditioned text feature $F_{IT}$. Finally, we reshape $F_{IT}$ to the image-like feature and a long residual path is added from $F_I$ to $F_{IT}$.

\begin{align}
    & A = Flat(Pool(Q)), A \in \mathcal{R}^{H_AW_A(n) \times C}, 
    \label{eq:cross_attn1} \\
    & Q = Flat(Q), Q \in \mathcal{R}^{HW(N) \times C}, 
    \label{eq:cross_attn2} \\
    & V_A = \sigma(A, K^T, V), V_A \in \mathcal{R}^{H_AW_A(n) \times C},
    \label{eq:cross_attn3} \\
    & F_{IT} = Reshape(\sigma(Q, A^T, V_A)) + F_I, \nonumber \\
    & F_{IT} \in \mathcal{R}^{H \times W \times C}
    \label{eq:cross_attn4}
\end{align}
where $\sigma$ denotes the softmax attention. Here, the computation complexity of agent cross attention is $O(Nnd) + O(NLd)$, which is smaller than vanilla cross attention's $O(NLd) + O(NLd)$, as $n \ll L$.

\subsection{Multiple Degradation Perception}
\label{subsec:mdp}

TransRFIR's encoder is utilized to dynamically encode the global context of the background and various degradation information in the degraded image. How can we verify if the encoder has successfully encoded all the degradation features? In contrast to the design of an implicit degradation context encoder proposed in \cite{potlapalli2023promptir}\cite{chen2023always}, we opt for a multi-task learning approach and design a module called Multiple Degradation Perception (MDP) to enhance the model's interpretability, which involves TransRFIR simultaneously recovering the image and performing multi-label classification on the encoded degradations by its encoder. This allows us to validate whether the model has successfully perceived different degradation features present in the image. 

MDP follows the last stage of HCEncoder and contains a Basic Convolution layer ($BC$) \cite{he2016deep}, a Global Pooling ($GP$) and three linear feedforward layers ($W_1$, $W_2$, $W_3$). We define the last stage feature of HCEncoder as $F_{l} \in \mathcal{R}^{H \times W \times C}$ while the output is $F_d \in \mathcal{R}^{1 \times D}$, where $D$ denotes the degradation number in our RFIR dataset. The whole process is shown in Equation \ref{eq:mdp1}, \ref{eq:mdp2} and \ref{eq:mdp3}.

\begin{align}
    & F_{MDP} = BC(F_l), F_{MDP} \in \mathcal{R}^{\frac{H}{2} \times \frac{W}{2} \times C},
    \label{eq:mdp1} \\
    & \hat{F}_{MDP} = GP(F_{MDP}), \hat{F}_{MDP} \in \mathcal{R}^{1 \times C},
    \label{eq:mdp2} \\
    & F_d = \hat{F}_{MDP} W_1W_2W_3, F_d \in \mathcal{R}^{1 \times D},
    \label{eq:mdp3}
\end{align}

\subsection{Training Objectives}
\label{subsec:training_obj}
The pipeline model we propose has two outputs: a multi-degradation category vector and the restored image, where the first prediction consists of a one-dimensional vector with five values (degradations), and the second is an image-like two-dimensional vector. We thus consider it a multi-task optimization problem. For the multi-degradation category classification, we use Binary Cross-Entropy ($BCE$) function to optimize multi-label classification, and for image restoration, we employ $L_1$ Loss. Empirically, we observe that these two values have losses of different magnitudes during optimization. Therefore, we balance the combination of these two losses based on homoscedastic uncertainty \cite{kendall2018multi} and formulate the training objectives as Equation \ref{eq:bce}, \ref{eq:l1} and \ref{eq:joint}.

\begin{align}
    & BCE(y, \hat{y}) = -\sum^C _{i=1} [y_i log(\hat{y}_i) + (1-y_i)log(1-\hat{y}_i)],
    \label{eq:bce} \\
    & L_1(I, \hat{I}) = \frac{1}{N} \sum^H_{i=1} \sum^W_{j=1} |I_{(i,j)} - \hat{I}_{(i,j)} |,
    \label{eq:l1} \\
    & L_{MT} = \frac{1}{2\sigma_1^2}BCE(y, \hat{y}) + \frac{1}{2\sigma_2^2}L_1(I, \hat{I}) \nonumber \\
    & \quad \quad \quad \quad + log\sigma_1^2 + log\sigma_2^2,
    \label{eq:joint}
\end{align}
where $y$ and $\hat{y}$ in Equation \ref{eq:bce} denote the ground truth of degradation categories in the degraded image and predicted categories. $C$ is the number of degradation categories. In Equation \ref{eq:l1}, $I$ and $\hat{I}$ are the ground truth image and predicted image. $H$ and $W$ denote the height and width of the image during training, and $N=H \times W$. In Equation \ref{eq:joint}, $\sigma_1$ and $\sigma_2$ are two learnable parameters to weigh and smooth two loss items, which alleviate the domination of optimization by the loss item with a large value.

\section{Experiments}
\label{sec:experiments}

\subsection{Dataset Settings}
The experiments include two parts. \textbf{Firstly}, we train models on our proposed RFIR dataset, including 107,396 samples for training, 15,342 samples for validation and 30,685 for testing. \textbf{Secondly}, to validate the generalization and effectiveness of our proposed architecture of TransRFIR, we introduce four common image restoration datasets for different single degradation removal, containing GoPro \cite{nah2017deep}, Rain100H \cite{yang2016joint}, LSRW \cite{hai2023r2rnet} and SOTS \cite{li2018benchmarking}.

\subsection{Model and Experimental Settings}
\textbf{Basic Models.} Besides our proposed TransRFIR, we select models of three paradigms for structure migration, experiment and comparison on proposed RFIR dataset, including task-agnostic models of global self-attention-based Restormer \cite{zamir2022restormer}, windows self-attention-based SwinIR \cite{liang2021swinir} and CNN-based NAFNet \cite{chen2022simple}, all-in-one models of PromptIR \cite{potlapalli2023promptir} and TransWeather (TWeather) \cite{valanarasu2022transweather}, and original text-driven model of IntructIR \cite{conde2024high}. For task-agnostic and all-in-one models, we migrate the paradigm of TransRFIR's text-driven mode to them. We do not choose models with multi-degradation-aware heads \cite{liu2022tape,li2020all,li2022all,chen2021pre} due to large memory and unflexible migration. Moreover, to validate the generalization of our proposed pipeline, we also evaluate prompt-free TransRFIR on four well-known image restoration benchmarks, including GoPro \cite{nah2017deep}, Rain100H \cite{yang2016joint}, LSRW \cite{hai2023r2rnet} and SOTS \cite{li2018benchmarking}. We include Restormer, NAFNet, InstructIR, DA-CLIP \cite{luo2023controlling} and Retinexformer \cite{cai2023retinexformer} for comparison.

\textbf{Cross Attention Methods.} Furthermore, to compare performances of image-text fusion, besides our proposed MHACA, we also include flash attention-based MHCA \cite{wu2023referring} and MHLCA \cite{choromanski2020rethinking} in experiments. We keep the same hyperparameters with our TransRFIR as Restormer on staged blocks \{4, 6, 6, 8\}, refinement blocks (4), attention heads \{1, 2, 4, 8\}, channel numbers \{48, 96, 192, 384\} and channel expansion ratio of GDFN (2.66). 

\textbf{Training.} We resize the image as 128 $\times$ 128 (px) and adopt random rotation and flip as the augmentation during the training. We train all models for 150 epochs with a learning rate interval of 0.001 to 0.01, accompanied by a cosine scheduler. We set the batch size as 16 and choose SGDM as the optimizer with a momentum of 0.937 and weight decay of 1e-4. Exponential Moving Average (EMA) is also included for model weight smoothing. We train all models on two RTX 3090ti GPUs. Moreover, for the generalization experiments, we train TransRFIR for 200 epochs with AdamW optimizer, whose weight decay is 1e-4. The batch size is set to 32 with the initial learning rate of 5e-4.

\textbf{Evaluation.} To evaluate the similarity between paired images from various perspectives, we use quantity metric PSNR \cite{wang2004image}, structural metric SSIM \cite{wang2004image} and learned perceptual metric LPIPS \cite{zhang2018unreasonable}. Furthermore, we adopt multi-label accuracy to evaluate the multi-degradation classification of MDP. Exactly, we identify that one sample is predicted correctly only based on each degradation category in the image is predicted correctly.

\begin{align}
    Acc = \frac{1}{N}\sum ^N_{i=1}\frac{\prod ^M_{j=1} y_{ij}}{M}
\end{align}
where $N$ represents the number of samples, each sample has $M$ labels, where $y_{ij}$ indicates whether the $j$-th label of the $i$ sample is predicted correctly (1-correct, 0-wrong).

\begin{table*}[h]
\scriptsize
  \setlength\tabcolsep{1.2pt}
  \caption{\textmd{Overall quantity experiments on RFIR dataset with models of various types under our proposed pipeline.}}
  \vspace{-1mm}
  \label{tab:rfir_exp}
  \begin{tabular}{c|ccc|ccc|ccc|ccc|ccc|ccc|ccc|c}
    \toprule
     & \multicolumn{18}{c}{\textbf{Number of Degradations}} \vline &  & \multirow{3}[2]{*}{\textbf{All}} & & \multirow{4}[1]{*}{\textbf{Accu}} \\
    \cmidrule(lr){2-19}
    \textbf{Basic} & \multicolumn{3}{c}{\textbf{One Degradation}} \vline & \multicolumn{6}{c}{\textbf{Double Degradations}} \vline & \multicolumn{9}{c}{\textbf{Triple Degradations}} \vline \\
     \cmidrule(lr){2-4} \cmidrule(lr){5-10} \cmidrule(lr){11-19} \cmidrule(lr){20-22}
    \textbf{Models} & \multicolumn{3}{c}{\textbf{1-1}} \vline & \multicolumn{3}{c}{\textbf{2-1}} \vline & \multicolumn{3}{c}{\textbf{2-2}} \vline & \multicolumn{3}{c}{\textbf{3-1}} \vline & \multicolumn{3}{c}{\textbf{3-2}} \vline & \multicolumn{3}{c}{\textbf{3-3}} \vline &  \multirow{2}[2]{*}{P} & \multirow{2}[2]{*}{S} & \multirow{2}[2]{*}{L} \\
     \cmidrule(lr){2-4} \cmidrule(lr){5-7} \cmidrule(lr){8-10} \cmidrule(lr){11-13} \cmidrule(lr){14-16} \cmidrule(lr){17-19} \cmidrule(lr){23-23} 
     &  P $\uparrow$ & S $\uparrow$ & L $\downarrow$ & P & S & L & P & S & L & P & S & L & P & S & L & P & S & L \\
     \midrule
     \multicolumn{22}{c}{\textbf{Single-Segment Restoration Prompt}} \\
     \midrule
     \multicolumn{22}{c}{Task-agnostic Models (under the pipeline of TransRFIR)} \\
     \midrule
     NAFNet & \uline{25.73} & \uline{0.868} & \uline{0.052} & 26.61 & 0.851 & 0.065 & 19.17 & 0.668 & 0.157 & 21.63 & 0.689 & 0.108 & 17.98 & 0.641 & 0.139 & 16.07 & 0.561 & 0.198 & 24.53 & 0.815 & \textbf{0.076} & 91.88\\
     SwinIR & 25.54 & 0.858 & 0.061 & 26.63 & 0.851 & \textbf{0.064} & 19.42 & \textbf{0.674} & 0.152 & 21.98 & 0.696 & 0.101 & 18.10 & 0.657 & 0.131 & 16.16 & 0.576 & 0.192 & 24.49 & 0.813 & 0.079 & 91.52 \\
     Restormer & 25.69 & 0.860 & 0.059 & \uline{26.69} & \uline{0.854} & 0.069 & \textbf{20.40} & \uline{0.671} & \uline{0.152} & \textbf{22.14} & \textbf{0.718} & 0.092 & 20.17 & 0.668 & 0.129 & 17.23 & 0.581 & 0.184 & \textbf{24.82} & \uline{0.816} & 0.078 & 92.15\\
     \midrule
     \multicolumn{22}{c}{All-In-One Models (under the pipeline of TransRFIR)} \\
     \midrule
     TWeather & 23.55 & 0.793 & 0.085 & 22.48 & 0.790 & 0.095 & 19.47 & 0.662 & 0.155 & 20.02 & 0.700 & \textbf{0.089} & \textbf{20.37} & \textbf{0.696} & 0.131 & \textbf{18.27} & \textbf{0.592} & 0.187 & 22.27 & 0.765 & 0.010 & 87.67\\
     PromptIR & 25.68 & 0.860 & 0.060 & 26.56 & 0.841 & 0.072 & 19.98 & 0.661 & \textbf{0.149} & 21.79 & 0.711 & 0.094 & 20.23 & \uline{0.679} & \textbf{0.128} & 17.88 & 0.584 & \uline{0.184} & 24.75 & 0.811 & 0.080 & \uline{92.79}\\
     \midrule
     \multicolumn{22}{c}{Text-Guided Models} \\
     \midrule
     InstructIR & \textbf{25.76} & \textbf{0.870} & \textbf{0.050} & 26.61 & 0.852 & 0.065 & 19.14 & 0.661 & 0.159 & 21.67 & 0.691 & 0.105 & 17.98 & 0.640 & 0.140 & 16.05 & 0.554 & 0.200 & 24.55 & 0.816 & 0.077 & 92.16\\
     \midrule
     \textbf{TransRFIR} & 25.66 & 0.856 & 0.058 & \textbf{26.70} & \textbf{0.861} & \uline{0.064} & \uline{20.03} & 0.670 & 0.153 & \uline{22.12} & \uline{0.717} & \uline{0.092} & \uline{20.23} & 0.671 & \uline{0.129} & \uline{18.10} & \uline{0.588} & \textbf{0.182} & \uline{24.81} & \textbf{0.817} & \uline{0.077} & \textbf{92.87}\\
     \midrule
     \multicolumn{22}{c}{\textbf{Double-Segment Restoration Prompt}} \\
     \midrule
     \multicolumn{22}{c}{Task-agnostic Models (under the pipeline of TransRFIR)} \\
     \midrule
     NAFNet & \uline{25.82} & \uline{0.868} & \textbf{0.054} & 26.67 & 0.853 & \uline{0.062} & 19.19 & 0.668 & 0.157 & 21.71 & 0.692 & 0.107 & 18.08 & 0.650 & 0.137 & 16.07 & 0.560 & 0.198 & 24.61 & 0.816 & \textbf{0.076} & 92.33 \\
     SwinIR & 25.71 & 0.862 & 0.059 & 26.67 & 0.855 & 0.064 & 19.43 & 0.673 & \uline{0.152} & 22.09 & 0.699 & 0.100 & 18.16 & 0.652 & 0.129 & 16.16 & 0.578 & 0.191 & 24.60 & 0.816 & 0.078 & 91.67\\
     Restormer & \textbf{26.17} & \textbf{0.871} & 0.056 & \uline{26.74} & \uline{0.859} & 0.063 & \textbf{20.44} & \textbf{0.675} & 0.152 & \uline{22.20} & 0.723 & 0.090 & 20.19 & 0.670 & 0.130 & 17.26 & 0.582 & 0.183 & \textbf{25.07} & \uline{0.824} & 0.075 & \uline{93.31}\\
     \midrule
     \multicolumn{22}{c}{All-In-One Models (under the pipeline of TransRFIR)} \\
     \midrule
     TWeather & 23.55 & 0.793 & 0.084 & 22.61 & 0.820 & 0.091 & 19.47 & 0.663 & 0.156 & 21.97 & \textbf{0.760} & \textbf{0.086} & \textbf{20.41} & \textbf{0.699} & 0.128 & \textbf{18.29} & \textbf{0.595} & 0.185 & 22.32 & 0.778 & 0.100 & 89.96\\
     PromptIR & 25.81 & 0.866 & 0.059 & 26.69 & 0.846 & 0.064 & 20.01 & 0.664 & 0.158 & 21.85 & 0.714 & \uline{0.089} & \uline{20.24} & \uline{0.679} & \textbf{0.126} & 17.89 & \uline{0.589} & \uline{0.182} & 24.86 & 0.817 & 0.077 & 93.20 \\
     \midrule
     \multicolumn{22}{c}{Text-Guided Models} \\
     \midrule
     InstructIR & 25.80 & 0.861 & \uline{0.056} & 26.70 & 0.859 & 0.063 & 19.14 & 0.662 & 0.157 & 21.71 & 0.698 & 0.101 & 18.03 & 0.644 & 0.138 & 16.08 & 0.554 & 0.199 & 24.60 & 0.814 & 0.077 & 92.31 \\
     \midrule
     \textbf{TransRFIR} & 25.97 & 0.868 & 0.058 & \textbf{26.79} & \textbf{0.868} & \textbf{0.061} & \uline{20.09} & \uline{0.674} & \textbf{0.151} & \textbf{22.21} & \uline{0.724} & 0.091 & 20.24 & 0.676 & \uline{0.127} & \uline{18.12} & 0.588 & \textbf{0.181} & \uline{24.99} & \textbf{0.826} & \uline{0.076} & \textbf{93.38}\\
    \bottomrule
  \end{tabular}
\end{table*}

\begin{table}
    \setlength\tabcolsep{1.0pt}
    \caption{\textmd{Comparison of different multi-head self-attention under features with different sizes.}}
    \vspace{-1mm}
    \label{tab:mhsa_compare}
    \centering
    \begin{tabular}{c|cc|ccc}
    \toprule
        \textbf{Methods} & \textbf{Params(M)}$\downarrow$ & \textbf{FLOPs(M)}$\downarrow$ & \textbf{PSNR}$\uparrow$ & \textbf{SSIM}$\uparrow$ & \textbf{LPIPS}$\downarrow$ \\
    \midrule
    \multicolumn{6}{c}{\textcolor{blue}{Stage 3: }\textbf{$D$=192, $N_h$=4, $H$=32, $W$=32}} \\
    \multicolumn{6}{c}{\textcolor{red}{Stage 4: }\textbf{$D$=384, $N_h$=8, $H$=16, $W$=16}} \\
    \midrule
    \multirow{2}[2]{*}{MHSA \cite{dao2022flashattention}}  & \textcolor{blue}{\uline{0.150}} & \textcolor{blue}{555.42} & \multirow{2}[2]{*}{25.06} & \multirow{2}[2]{*}{\uline{0.826}} & \multirow{2}[2]{*}{0.076}\\
                                      & \textcolor{red}{\uline{0.595}} & \textcolor{red}{\textbf{202.21}} \\
    \midrule
    \multirow{2}[2]{*}{MDTA \cite{zamir2022restormer}}  & \textcolor{blue}{0.153} & \textcolor{blue}{312.61} & \multirow{2}[2]{*}{\uline{25.07}} & \multirow{2}[2]{*}{0.824} & \multirow{2}[2]{*}{\textbf{0.075}} \\
                                      & \textcolor{red}{0.600} & \textcolor{red}{307.30} \\
    \midrule
    \multirow{2}[2]{*}{MHLSA \cite{choromanski2020rethinking}}  & \textcolor{blue}{0.197} & \textcolor{blue}{402.65} & \multirow{2}[2]{*}{24.53} & \multirow{2}[2]{*}{0.822} & \multirow{2}[2]{*}{0.070} \\
                                      & \textcolor{red}{0.787} & \textcolor{red}{402.65} \\
    \midrule
    \multicolumn{6}{c}{Agent Size = (7 $\times$ 7)} \\
    \midrule
    \multirow{2}[2]{*}{\textbf{MHASA}} & \textcolor{blue}{\textbf{0.150}} & \textcolor{blue}{\uline{306.20}} & \multirow{2}[2]{*}{24.99} & \multirow{2}[2]{*}{0.817} & \multirow{2}[2]{*}{0.076} \\
                                       & \textcolor{red}{\textbf{0.594}} & \textcolor{red}{\textbf{304.14}} \\
    \midrule
    \multicolumn{6}{c}{Agent Size = (8 $\times$ 8)} \\
    \midrule
    \multirow{2}[2]{*}{\textbf{MHASA}} & \textcolor{blue}{\textbf{0.150}} & \textcolor{blue}{306.24} & \multirow{2}[2]{*}{25.01} & \multirow{2}[2]{*}{0.818} & \multirow{2}[2]{*}{0.076} \\
                                       & \textcolor{red}{\textbf{0.594}} & \textcolor{red}{\uline{304.18}} \\
    \midrule
    \multicolumn{6}{c}{Agent Size = (12 $\times$ 12)} \\
    \midrule
    \multirow{2}[2]{*}{\textbf{MHASA}} & \textcolor{blue}{\textbf{0.150}} & \textcolor{blue}{306.34} & \multirow{2}[2]{*}{\textbf{25.08}} & \multirow{2}[2]{*}{\textbf{0.827}} & \multirow{2}[2]{*}{\uline{0.075}} \\
                                       & \textcolor{red}{\textbf{0.594}} & \textcolor{red}{\uline{304.20}} \\
    \bottomrule
    \end{tabular}
\end{table}

\subsection{Quantitive Results}

\begin{figure}
    \includegraphics[width=0.99\linewidth]{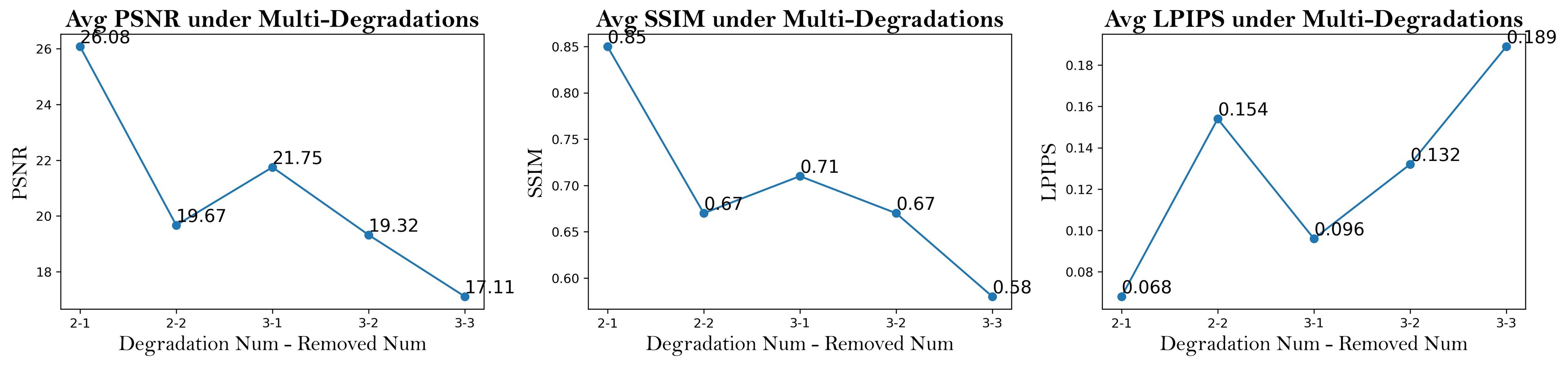}
    \vspace{-3mm}
    \caption{\textmd{Average metrics of all models under multiple degradations.}}
    \label{fig:line_chart}
\end{figure}

\begin{figure}
    \includegraphics[width=0.99\linewidth]{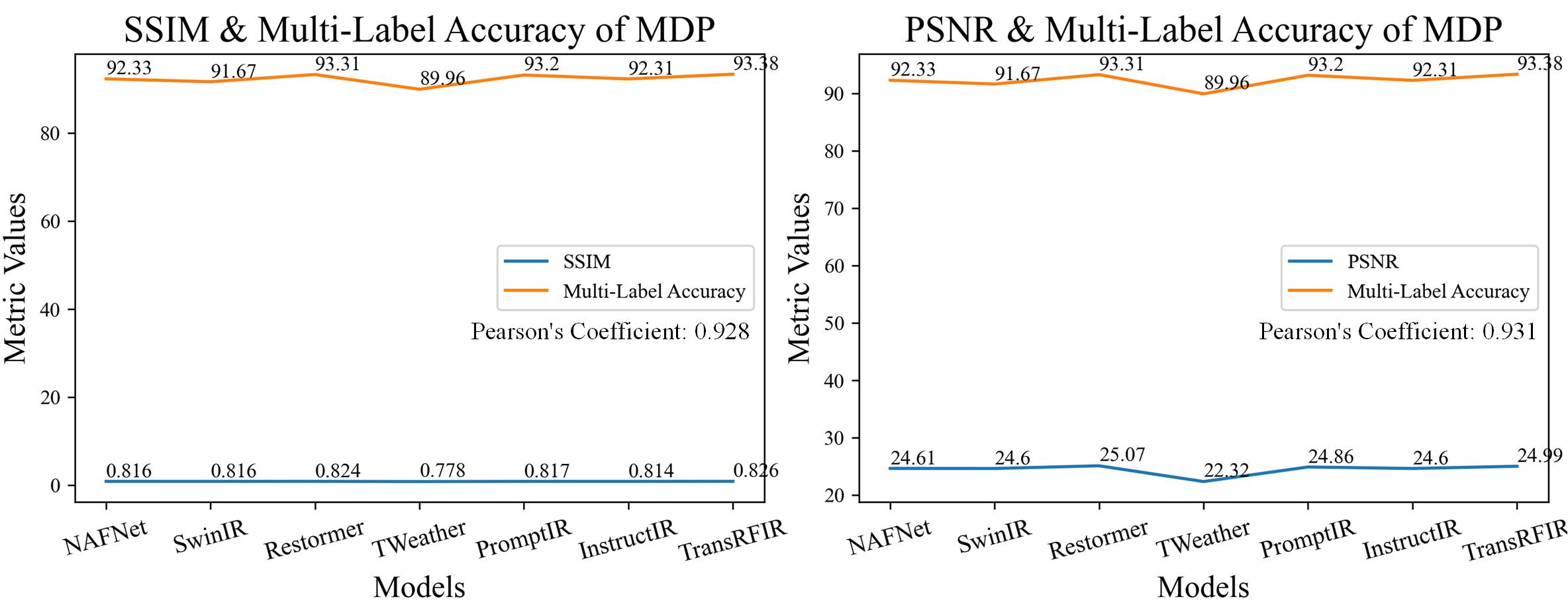}
    \vspace{-3mm}
    \caption{\textmd{Correlation between metrics of image restoration and multi-degradation perception.}}
    \label{fig:correlation}
\end{figure}

\begin{table}
    \setlength\tabcolsep{0.7pt}
    \caption{\textmd{Comparison of fusion methods on multi-scale feature maps.}}
    \vspace{-1mm}
    \centering
    \begin{tabular}{c|cc|ccc}
    \toprule
    \textbf{Methods} & \textbf{Params(M)}$\downarrow$ & \textbf{FLOPs(M)}$\downarrow$ & \textbf{PSNR}$\uparrow$ & \textbf{SSIM}$\uparrow$ & \textbf{LPIPS}$\downarrow$ \\
    \midrule
    \multicolumn{6}{c}{\textbf{TransRFIR [$F_I$: (1$\times$384$\times$16$\times$16), $F_T$: (1$\times$20$\times$384)]}} \\
    \midrule
    MHCA \cite{wu2023referring} & \uline{0.591} & 85.33 & \uline{25.01} & \uline{0.826} & 0.077 \\
    MHLCA \cite{choromanski2020rethinking} & 0.787 & 217.06 & 24.52 & 0.820 & 0.085 \\
    \midrule
    \multicolumn{6}{c}{Agent Size = (7 $\times$ 7)} \\
    \midrule
    \textbf{MHACA} & \textbf{0.297} & \textbf{12.77} & 24.99 & 0.826 & \uline{0.076} \\
    \midrule
    \multicolumn{6}{c}{Agent Size = (12 $\times$ 12)} \\
    \midrule
    \textbf{MHACA} & \textbf{0.297} & \uline{12.80} & \textbf{25.08} & \textbf{0.827} & \textbf{0.075} \\
    \midrule
    \multicolumn{6}{c}{\textbf{NAFNet [$F_I$: (1$\times$512$\times$8$\times$8), $F_T$: (1$\times$20$\times$512)]}} \\
    \midrule
    MHCA & 1.051 & \uline{45.35} & 24.55 & \uline{0.816} & \uline{0.076} \\
    MHLCA & \uline{1.049} & 88.08 & 24.30 & 0.810 & 0.079 \\
    \midrule
    \multicolumn{6}{c}{Agent Size = (7 $\times$ 7)} \\
    \midrule
    \textbf{MHACA} & \textbf{0.527} & \textbf{21.33} & 24.53 & 0.815 & 0.076 \\
    \midrule
    \multicolumn{6}{c}{Agent Size = (12 $\times$ 12)} \\
    \midrule
    \textbf{MHACA} & \textbf{0.527} & \uline{21.38} & \textbf{24.56} & \textbf{0.817} & \textbf{0.075} \\
    \bottomrule
    \end{tabular}
    \label{tab:fusion_methods}
\end{table}

\textbf{Overall Performances.} Table \ref{tab:rfir} presents the overall performances of different models under our proposed pipeline as Figure \ref{fig:teaser} presents. We categorize test set into three major groups and six subgroups, which encompass single, double, and triple degradation, with the latter two including two and three variations, respectively. Similarly, our models are also classified into three categories. \textbf{Firstly}, as Figure \ref{fig:line_chart} shows, no matter the number of degradation in the degraded image, regardless of the models, the improvement obtained by removing single degradation is much greater than that achieved by removing double and triple degradations. This directly demonstrates the complexity of spatial contextual patterns corresponding to multiple degradations, making their recognition and removal much more challenging. \textbf{Secondly}, when we adopt double-segment restoration prompts, a slight promotion occurs compared with using single-segment restoration prompts. This demonstrates that when providing external strong prior knowledge to the model, the model can to some extent understand and reflect this in the results of interference recovery. \textbf{Thirdly}, transformer-based models present overall advancement compared with CNN-based models, which indicates the significance of dynamic input-dependency during restoration. \textbf{Fourthly},  we find that our pipeline can be migrated to task-agnostic and all-in-one models seamlessly and adapt well, achieving superior performance compared to the native text-guided model InstructIR. Fifthly, as Figure \ref{fig:correlation} presents, the metrics of image restoration, including PNSR and SSIM, have highly correlation with the multi-label accuracy of multi-degradation perception, where the Pearson's coefficients are above 0.90 both, proving the rationality and effectiveness of our proposed multi-task pipeline.

\textbf{Various Self-Attention.} As Table \ref{tab:mhsa_compare} presents, our proposed MHASA achieves state-of-the-art performances compared with other multi-head self-attention, including flash attention-based vanilla Multi-Head Self-Attention (MHSA), Multi-Dconv Head Transposed Attention (MDTA), Multi-Head Linear Self-Attention (MHLSA). For the three metrics of image restoration, the performances of MHASA become better when the agent size is larger, which also leads to the slight promotion of FLOPs. Exactly, the performance of MHASA is still worse than the other three self-attention methods when the agent size is below $8 \times 8$, but things change when the agent size becomes $12 \times 12$. More importantly, our MHASA has relatively lower parameters and FLOPs compared with other methods for feature maps of different sizes. Nevertheless, flash attention-based MHSA still obtains the lowest FLOPs on the feature map of stage 4. In a word, our MHASA obtains the tradeoff between accuracy and complexity.

\begin{figure*}
    \includegraphics[width=0.99\linewidth]{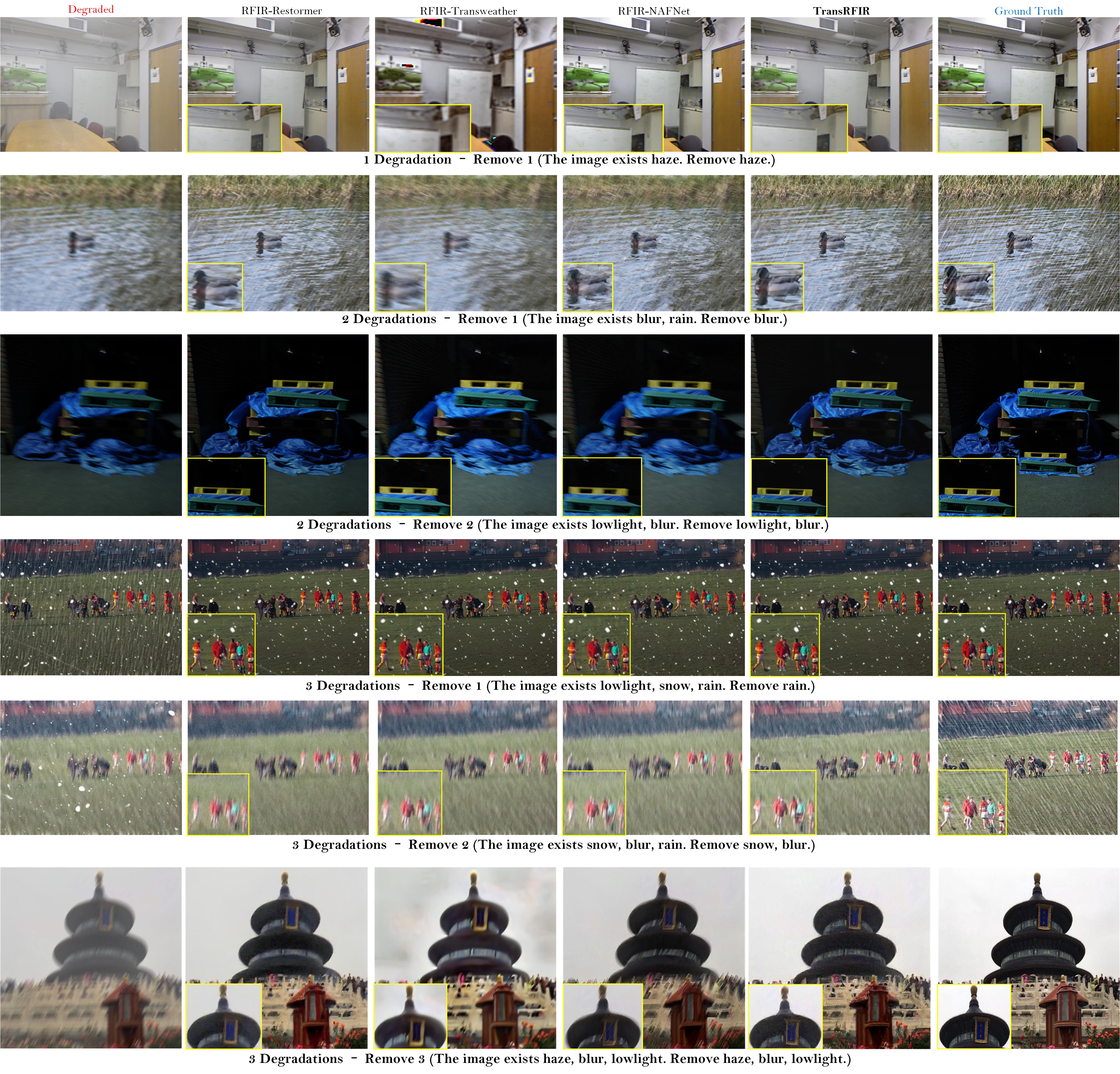}
    \vspace{-3mm}
    \caption{\textmd{Overall visualization by various model predictions on degradations with different types.}}
    \label{fig:initial_prediction}
\end{figure*}

\textbf{Various Cross-Attention.} As an essential module to convey the prompt signal to image restoration decoder, cross attention fuses the textual prompt containing dense information with degraded image from the perspective of global context by calculating the similarity between image representations and text embeddings. Table \ref{tab:fusion_methods} presents the performances of various cross attention methods on two models with different structures. \textbf{Firstly}, for feature maps of two sizes, our MHACA possesses much lower parameters and FLOPs among all methods with a relatively low increasing rate on FLOPs also, which proves the scalability of MHACA. \textbf{Secondly}, when the agent size is set $7 \times 7$, the performance of MHACA is better than MHLCA but still falls behind MHCA. Nevertheless, when the agent size is set $12 \times 12$, our MHACA obtains the state-of-the-art performances on three different restoration metrics.

\subsection{Qualitative Results}

\begin{figure*}
    \includegraphics[width=1.00\linewidth]{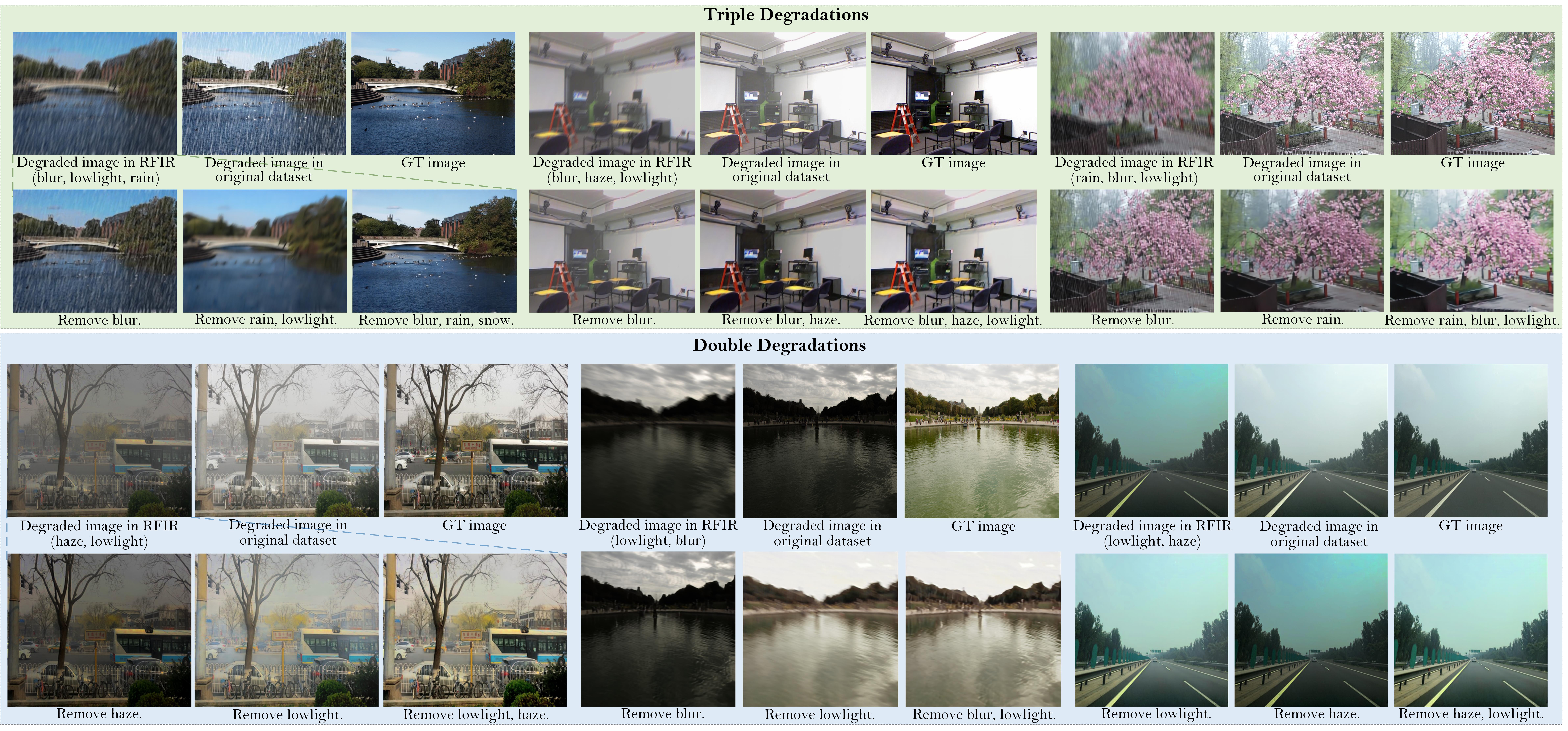}
    \vspace{-5mm}
    \caption{\textmd{Prediction samples with different restoration prompts for Out-Of-Distribution (OOD) test by TransRFIR.}}
    \label{fig:general_test}
\end{figure*}

\begin{figure*}
    \includegraphics[width=1.00\linewidth]{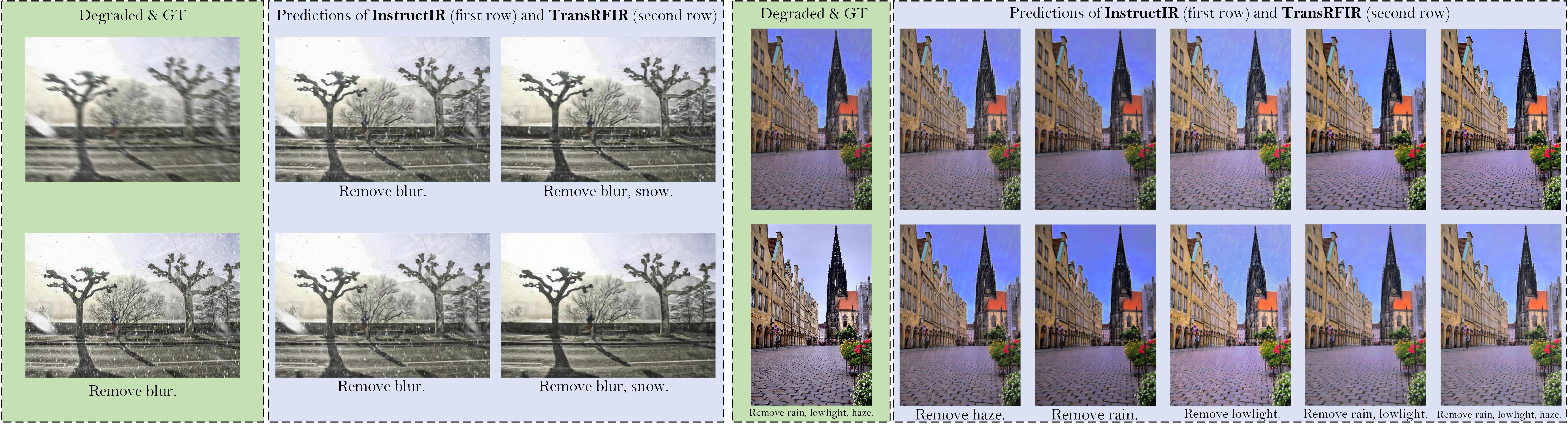}
    \vspace{-5mm}
    \caption{\textmd{Comparison between CNN-based InstructIR and Transformer-based TransRFIR on multi-degradations.}}
    \label{fig:instruct_compare}
\end{figure*}

\textbf{Overall visualization by various models.} Figure \ref{fig:initial_prediction} shows the predictions by various models on our RFIR dataset. \textbf{Firstly}, when we focus on the first row, all models perform nicely for single degradation removal based on the image with one degradation. \textbf{Secondly}, when we look at the second and third rows, which exist two degradations in the image, we first compare the performances between various models, our TransRFIR achieves satisfactory restoration results, as perceived by human vision, whether it is the removal of blur in the second row or the removal of low light and blur in the third row. Furthermore, for the simultaneous removal of blur and low light degradation in the third row, transformer-based models outperform the CNN-based NAFNet. This reflects the superior ability of self-attention in understanding and recognizing complex contexts compared to fixed-weight convolutions. \textbf{Thirdly}, as the degradation increases further, we observe the last three rows of images. When only a single degradation is removed, all models achieve similar satisfactory results. However, as the degradation becomes double or triple, we observe that the models are able to remove the corresponding interference to some extent based on the given prompt, but they do not achieve very satisfactory results from a human visual perspective. This reflects the difficulty for the models to recognize complex contextual patterns, especially in the case of triple degradation. 

\textbf{Generalization analysis for unseen prompts in the dataset.} As Figure \ref{fig:general_test} shows, we test our TransRFIR on data samples with unseen restoration prompts. \textbf{Firstly}, for images with triple degradation combinations, we input three different text prompts for removing specific degradation(s). For the first sample, when we command the model to remove blur, TransRFIR correctly recognizes the blur, removes it and keeps the degradations of rain and lowlight. When we input the prompt "Remove rain, lowlight.", we can see a bit of brightness enhancement on the restored image and the rain is also removed. When we desire to remove all degradations, we can see that our TransRFIR completes the tough job to a nice extent. For the other two samples, our TransRFIR also exhibits excellent generalization. \textbf{Secondly}, for the image with double degradations, in the first sample, when we only want to remove haze, judging from the reflection intensity of the objects in the figure, it seems that haze has been removed to some extent by TransRFIR. When we desire to remove low light, it is obvious that TransRFIR nicely removes the low light and maintains the haze, but the contrast does not seem to be quite the same as the original image, which is also a small problem. When we command TransRFIR to remove all degradations, TransRFIR does an overall good job of removing the low light and haze simultaneously, where the haze is not removed thoroughly. For the other two samples, TransRFIR also presents the nice generalization when facing other degradation combinations.

\textbf{Comparison between InstructIR and TransRFIR.} Figure \ref{fig:instruct_compare} presents the prediction results of CNN-based InstructIR and Transformer-based TransRFIR on a degraded image with double and triple degradations. For the sample with double degradations, we can observe our TransRFIR can remove blur, and both blur and snow degradations, but InstructIR does not do well on removing snow, there are still a few snowflakes left. For the sample with triple degradations, we notice that both models achieve good results; however, in comparison to the ground truth from a human visual perspective, TransRFIR appears to be closer in terms of brightness and clarity.

\subsection{Ablation Experiments}
To validate the effectiveness of our proposed pipeline, we conduct the ablation experiments as Table \ref{tab:Ablation} shows, \textbf{firstly}, when we adopt the one-stage guidance decoder, we observe a bit drop on all metrics. \textbf{Secondly}, when we stop using MDP for multi-degradation classification, the drop also happens, especially on LPIPS. \textbf{Thirdly}, we observe that the position encodings of query (image) and key (text prompt) in MHACA are also essential. \textbf{Fourthly}, we conduct experiments on RNN-based and transformer-based text encoders, including LSTM \cite{lstm}, BERT \cite{kenton2019bert} and RoBERTa \cite{liu2019roberta}, where we use GloVe \cite{pennington2014glove} as the text embedding for RNN-based encoders. It can be seen that LSTM has a large performance gap between two transformer-based text encoders.

\begin{table}
    \setlength\tabcolsep{0.4pt}
    \caption{\textmd{Ablation comparison of the proposed pipeline.}}
    \vspace{-1mm}
    \centering
    \begin{tabular}{c|ccc|ccc|ccc}
    \toprule
     \multirow{3}[2]{*}{\textbf{Methods}} & \multicolumn{3}{c}{\textbf{One Deg}} & \multicolumn{3}{c}{\textbf{Two Deg}} & \multicolumn{3}{c}{\textbf{Three Deg}}\\
     \cmidrule(lr){2-4} \cmidrule(lr){5-7} \cmidrule(lr){8-10} 
     &  \textbf{P}$\uparrow$ & \textbf{S}$\uparrow$ & \textbf{L}$\downarrow$ & \textbf{P} & \textbf{S} & \textbf{L} & \textbf{P} & \textbf{S} & \textbf{L} \\
    \midrule
       TransRFIR & 25.97 & 0.868 & 0.058 & 25.45 & 0.829 & 0.079 & 20.60 & 0.678 & 0.123\\
       \midrule
       One stage & 25.89 & 0.861 & 0.061 & 25.36 & 0.824 & 0.084 & 20.54 & 0.673 & 0.127 \\
       -MDP  & \textcolor{red}{\uline{25.86}} & \textcolor{red}{\uline{0.858}} & 0.060 & \textcolor{red}{\textbf{25.25}} & \textcolor{red}{\textbf{0.816}} & \textcolor{red}{\textbf{0.089}} & \textcolor{red}{\textbf{20.30}} & \textcolor{red}{\textbf{0.659}} & \textcolor{red}{\textbf{0.134}} \\
       -QueryPos & \textcolor{red}{\textbf{25.85}} & \textcolor{red}{\textbf{0.858}} & \textcolor{red}{\textbf{0.060}} & \textcolor{red}{\uline{25.34}} & \textcolor{red}{\uline{0.823}} & \textcolor{red}{\uline{0.084}} & \textcolor{red}{\uline{20.52}} & \textcolor{red}{\uline{0.671}} & \textcolor{red}{\uline{0.128}} \\
       -KeyPos & 25.93 & 0.864 & 0.058 & 25.43 & 0.828 & 0.079 & 20.55 & 0.673 & 0.124 \\
       \midrule
       LSTM & 21.42 & 0.708 & 0.081 & 20.23 & 0.689 & 0.102 & 16.72 & 0.521 & 0.155 \\
       BERT & \uline{25.96} & \uline{0.863} & \uline{0.059} & \uline{25.43} & \uline{0.828} & \uline{0.080} & \textbf{20.61} & \uline{0.677} & \textbf{0.122} \\
       RoBERTa & \textbf{25.99} & \textbf{0.870} & \textbf{0.057} & \textbf{25.45} & \textbf{0.828} & \textbf{0.078} & \uline{20.60} & \textbf{0.679} & \uline{0.124} \\
    \bottomrule
    \end{tabular}
    \label{tab:Ablation}
\end{table}

\begin{table}
    \setlength\tabcolsep{0.5pt}
    \caption{\textmd{Generalization experiments on image restoration datasets with different degradations.}}
    \vspace{-1mm}
    \centering
    \begin{tabular}{c|cc|cc|cc|cc}
    \toprule
    \multirow{3}[2]{*}{\textbf{Methods}}  & \multicolumn{2}{c}{\textbf{Deblurring}} & \multicolumn{2}{c}{\textbf{Deraining}} & \multicolumn{2}{c}{\textbf{Lowlight Enh}} & \multicolumn{2}{c}{\textbf{Dehazing}}\\
     & \multicolumn{2}{c}{GoPro} &  \multicolumn{2}{c}{Rain100H} & \multicolumn{2}{c}{LSRW} & \multicolumn{2}{c}{SOTS} \\
     \cmidrule(lr){2-3} \cmidrule(lr){4-5} \cmidrule(lr){6-7} \cmidrule(lr){8-9}
     & P & S & P & S & P & S & P & S \\
    \midrule
    NAFNet & 29.87 & 0.887 & 35.56 & 0.967 & 20.49 & 0.809 & 25.23 & 0.939\\
    Restormer & \textbf{31.82} & \textbf{0.918} & 34.81 & 0.962 & 20.41 & 0.806 & 24.09 & 0.927\\
    DA-CLIP & 30.88 & 0.903 & 33.91 & 0.926 & 20.26 & 0.802 & 24.69 & 0.932\\
    Retinexformer & - & - & - & - & 20.68 & 0.811 & - & -\\
    InstructIR & 29.93 & 0.892 & \textbf{36.84} & \textbf{0.973} & \uline{20.70} & \uline{0.820} & 25.20 & 0.938 \\
    \midrule
    \multicolumn{9}{c}{Agent Size of MHASA = 7 $\times$ 7} \\
    \midrule
    \textbf{TransRFIR} & 30.95 & 0.907 & 35.98 & 0.968 & 20.67 & 0.811 & \uline{25.24} & \uline{0.939} \\
    \midrule
    \multicolumn{9}{c}{Agent Size of MHASA = 12 $\times$ 12} \\
    \midrule
    \textbf{TransRFIR} & \uline{31.52} & \uline{0.912} & \uline{36.55} & \uline{0.971} & \textbf{20.73} & \textbf{0.822} & \textbf{25.30} & \textbf{0.943} \\
    \bottomrule
    \end{tabular}
    \label{tab:general_exp}
\end{table}

\subsection{Generalization Experiments}
\begin{figure}
    \includegraphics[width=1.00\linewidth]{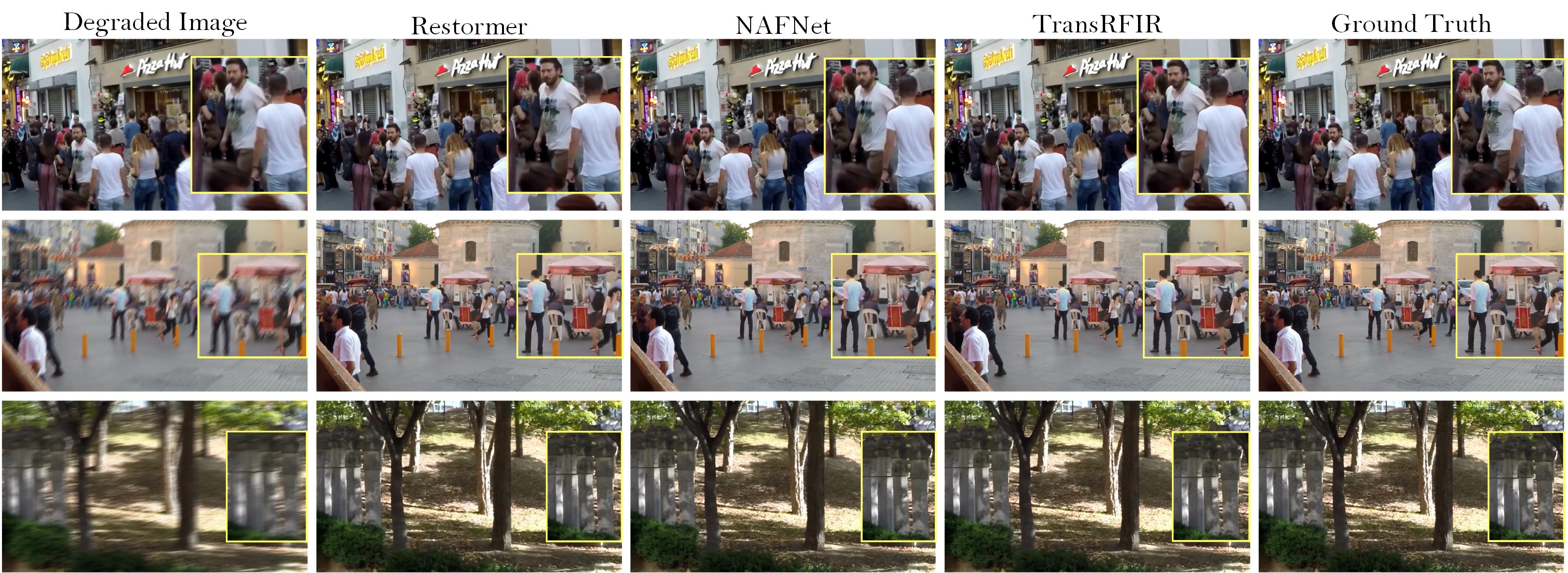}
    \vspace{-5mm}
    \caption{\textmd{Comparison of models on GoPro \cite{nah2017deep}.}}
    \label{fig:deblur_compare}
\end{figure}

\begin{figure}
    \includegraphics[width=1.00\linewidth]{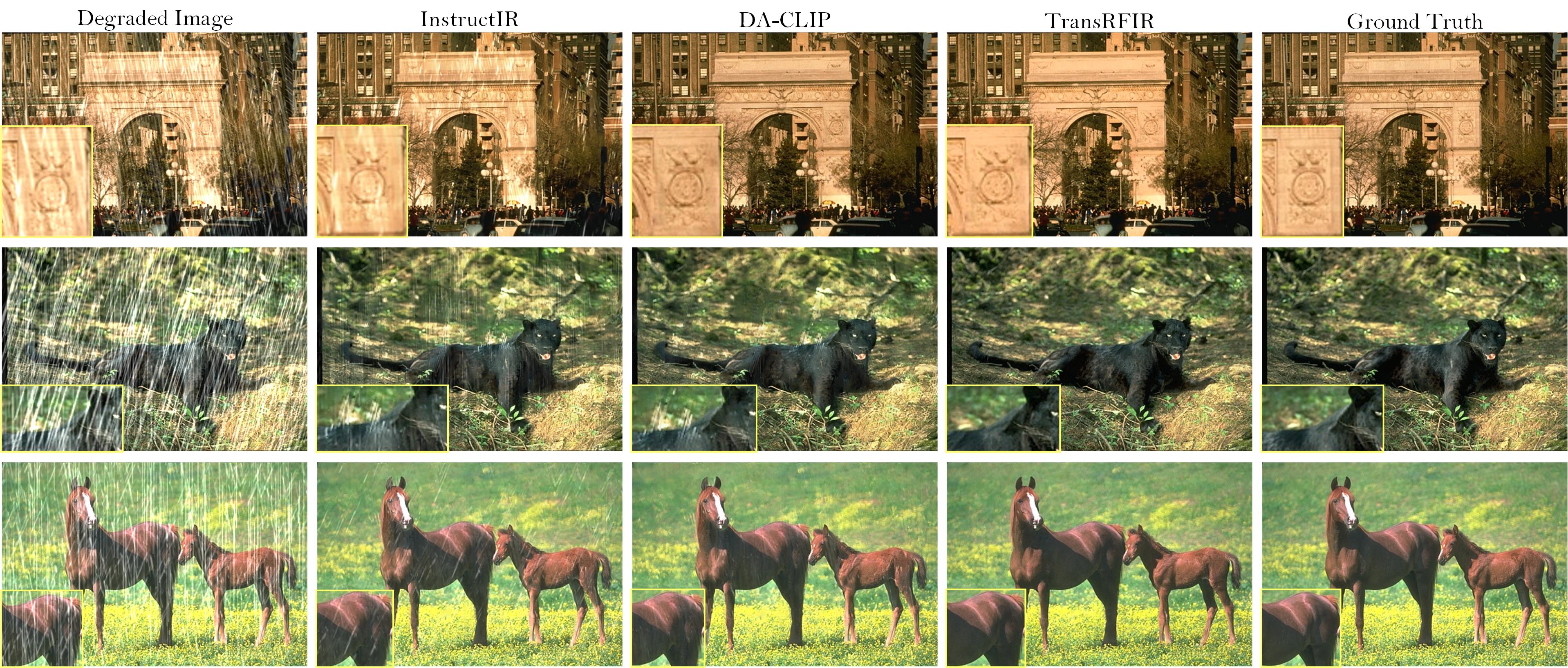}
    \vspace{-5mm}
    \caption{\textmd{Comparison of models on Rain100H \cite{yang2016joint}.}}
    \label{fig:derain_compare}
\end{figure}

\begin{figure}
    \includegraphics[width=1.00\linewidth]{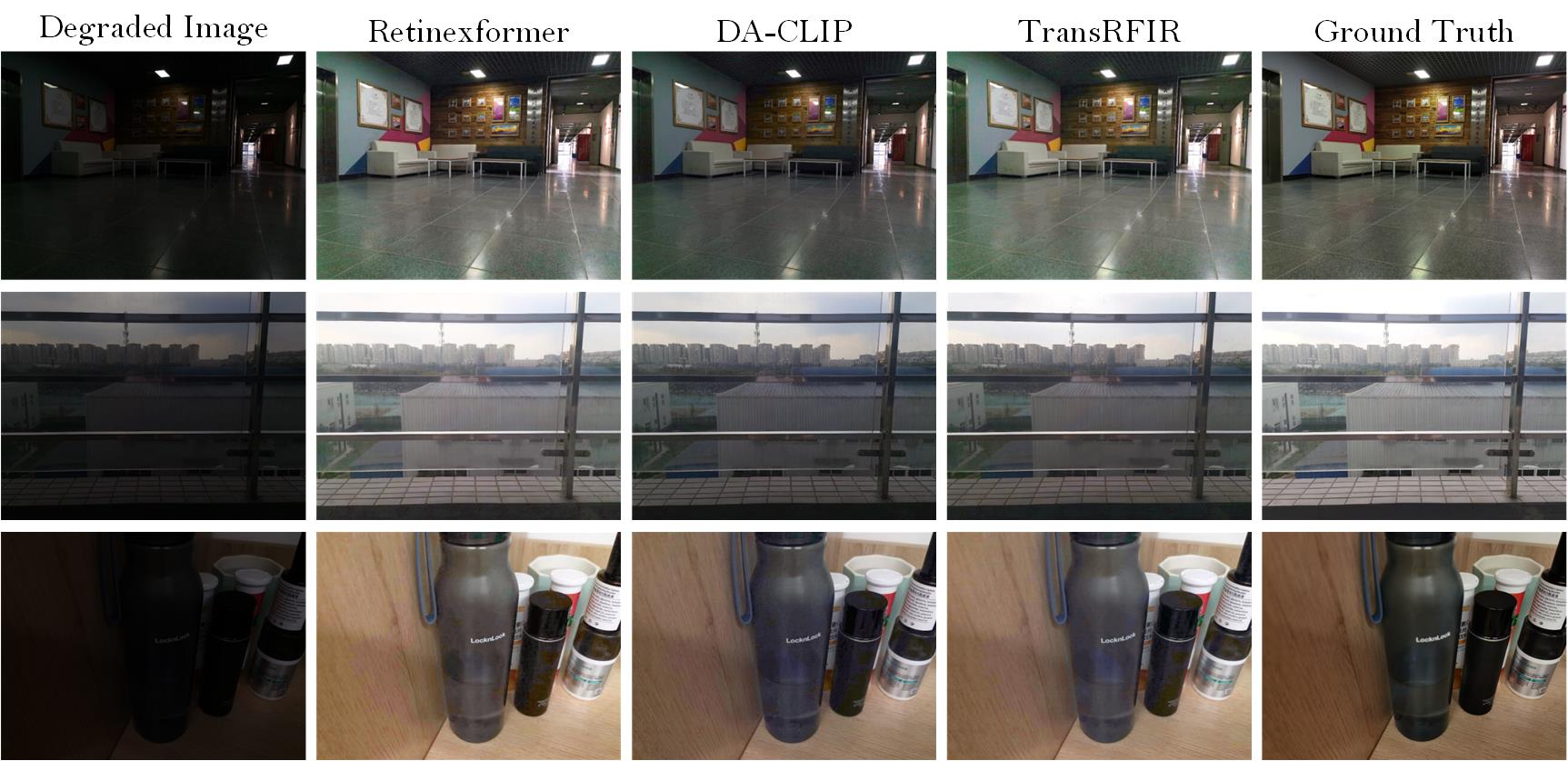}
    \vspace{-5mm}
    \caption{\textmd{Comparison of models on LSRW \cite{hai2023r2rnet}.}}
    \label{fig:delow_compare}
\end{figure}

\begin{figure}
    \includegraphics[width=1.00\linewidth]{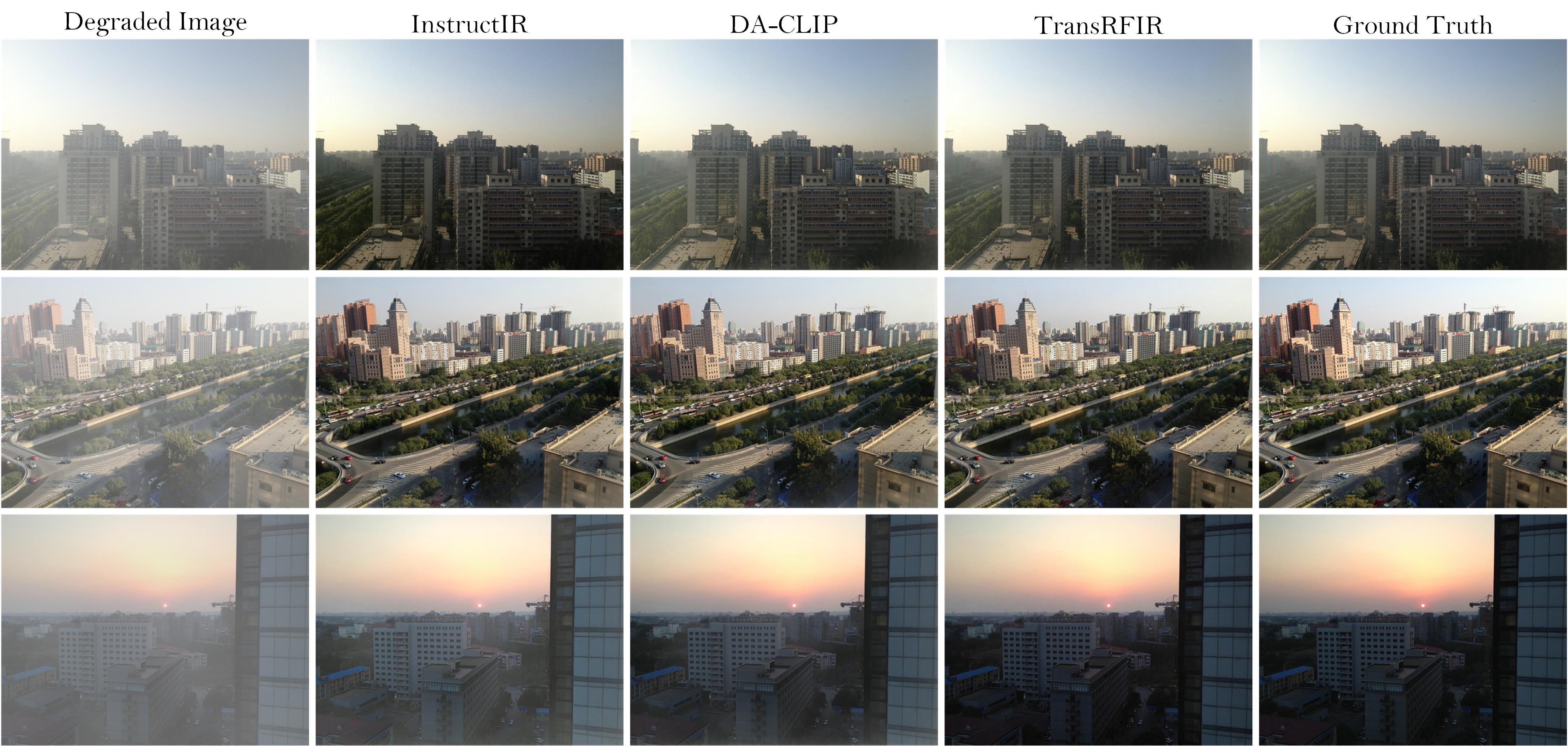}
    \vspace{-5mm}
    \caption{\textmd{Comparison of models on SOTS \cite{li2018benchmarking}.}}
    \label{fig:dehaze_compare}
\end{figure}

To further validate the effectiveness of our proposed pipeline on normal image restoration datasets, we compare our TransRFIR with other state-of-the-art methods on four datasets with different degradations. Table \ref{tab:general_exp} shows the quantitative results of experiments on four datasets while Figure \ref{fig:deblur_compare}, \ref{fig:derain_compare}, \ref{fig:delow_compare} and \ref{fig:dehaze_compare} show the qualitative results, respectively. Based on these, we find our TransRFIR achieves competitive performances with other state-of-the-art models. More importantly, TransRFIR obtains balanced performances on four benchmarks, which proves the generalization and effectiveness of our proposed pipeline.

\section{Conclusion}
\label{sec:conclusions}
In this paper, we propose a novel and highly challenging task called Referring Flexible Image Restoration (RFIR), where a large dataset of the same name is proposed. RFIR is proposed for tackling the specific degradation removal based on the text prompt. To tackle the challenge, we propose a corresponding pipeline called TransRFIR, which is a multi-task paradigm to restore image upon text prompt and recognize degradation types simultaneously. In TransRFIR, two attention modules, including MHASA and MHACA are devised. MHASA and MHACA are two efficient modules for self-attention feature modeling and feature fusion, respectively, which have linear complexity and relatively small parameters compared with their counterparts. Comprehensive experiments exhibit their generalization and advancement.

\section{Discussion}
\label{sec:discussion}

\textbf{A little story.} One day, while I was in the monitoring room of my grandmother's vegetable greenhouse, I was checking the status of the sprinkler system and the vegetables being sprayed in the vegetable field through the camera. On that foggy day, the wind was a little strong, causing the surveillance camera to occasionally shake. For the surveillance personnel, there are two types of image degradations they want to eliminate: fog and motion blur. Nevertheless, for the image restoration models, especially the all-in-one models, if there is no human intention as a supervisory signal, they will remove all degradations they perceive, including fog, blur, and the waterdrop sprayed by the sprinkler system. However, obviously, waterdrops should not be treated as degradation. Inspired by this, this work was born. In the future, We hope to implement a mature RFIR system in this vegetable greenhouse area.

\begin{figure}
    \includegraphics[width=1.00\linewidth]{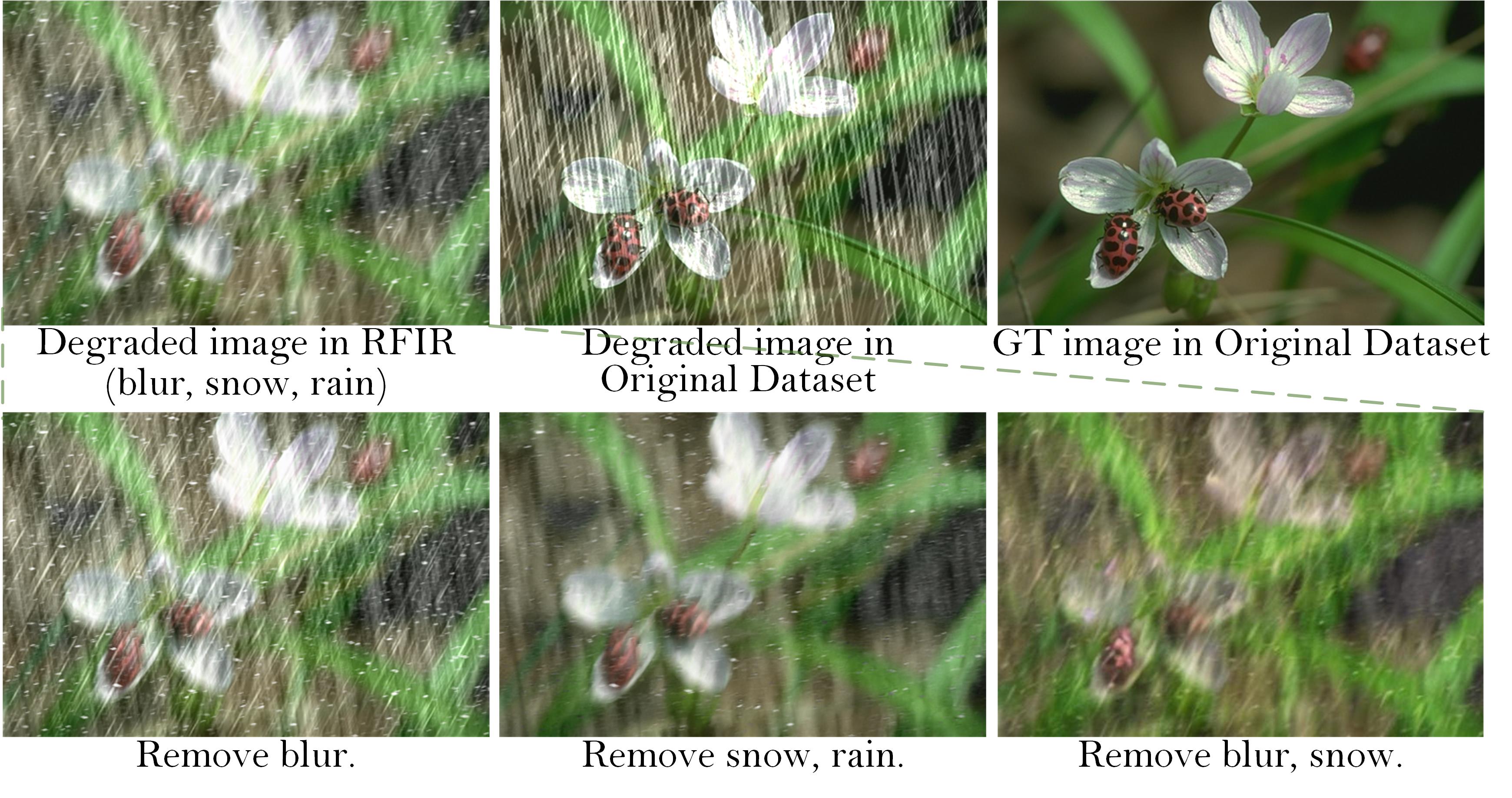}
    \vspace{-5mm}
    \caption{\textmd{Fail sample on combination of snow and rain.}}
    \label{fig:fail_samples}
\end{figure}

\textbf{Limitations.} \textbf{Firstly}, our proposed pipeline is oriented at U-Net-based image restoration networks, but as we all know, in addition to U-Net, GAN-based and Diffusion-based networks are also well-performed on image restoration tasks. \textbf{Secondly}, our proposed RFIR dataset are synthetic, although it is proved that it can be used for research on prompt-guided specific degradation removal, we still desire to collect some sample under real scenarios, which will be realized in the future. \textbf{Thirdly}, as Figure \ref{fig:fail_samples} presents, we notice when the rain and snow co-exist, the restoration result is not satisfactory, which is going to be a big problem to solve in the future. The possible reason is that there are some occlusions between raindrops and snowflake, which cause a tough for model to identify the exact shape of unreferred degradation. 

\textbf{Challenges.} \textbf{Firstly}, it is obvious that RFIR dataset contains various degradation patterns, which is challenging for information capacity of deep learning models. \textbf{Secondly}, how to maximize inter-domain gap between different multi-degradation matters. \textbf{Thirdly}, we notice that for some degradation combinations, generative models may be a better choice, as pixel-wise occlusion between different degradations exist in some samples, such as raindrops and snow. Nevertheless, how to design a specific generative model is a challenge. 

\section*{Acknowledgment}
This work is partially supported by the XJTLU AI University Research Centre and Jiangsu Province Engineering Research Centre of Data Science and Cognitive Computation at XJTLU. Also, it is partially funded by the Suzhou Municipal Key Laboratory for Intelligent Virtual Engineering (SZS2022004) as well as funding: XJTLU-REF-21-01-002, XJTLU- RDF-22-01-062, and XJTLU Key Program Special Fund (KSF-A-17).
This work received financial support from Jiangsu Industrial Technology Research Institute (JITRI) and Wuxi National Hi-Tech District (WND).

\printcredits

\bibliographystyle{cas-model2-names}

\bibliography{cas-dc-template}

\end{document}